\documentclass[preprint,12pt]{elsarticle}

\usepackage{amssymb}

\usepackage[colorlinks, linkcolor=blue, anchorcolor=blue, citecolor=blue]{hyperref}
\usepackage{algpseudocode} 
\usepackage{amsmath}
\usepackage{graphicx}
\usepackage{color}
\usepackage{booktabs}
\usepackage{hhline}
\usepackage[table]{xcolor}
\usepackage{bm}
\usepackage{mathtools}
\usepackage{enumitem}
\usepackage[flushleft]{threeparttable}
\usepackage{amsmath,latexsym,amssymb,amsthm,array,amsfonts,algorithm,algpseudocode,booktabs,graphicx,subfigure,multirow,cuted,stfloats}
\usepackage[section]{placeins}
\usepackage{geometry}
\usepackage[figuresright]{rotating}  % 旋转表格

\journal{Knowledge-Based Systems}

\begin{document}

\begin{frontmatter}



\title{An Edge-Aware Graph Autoencoder Trained on Scale-Imbalanced Data for Traveling Salesman Problems}


\author[inst1]{Shiqing Liu}

\affiliation[inst1]{organization={Nature Inspired Computing and Engineering, Faculty of Technology, Bielefeld University},%Department and Organization 
            city={Bielefeld},
            postcode={33619}, 
            country={Germany}}

\author[inst2]{Xueming Yan}
\author[inst1,inst3]{Yaochu Jin}

\affiliation[inst2]{organization={School of Information Science and Technology, Guangdong University of Foreign Studies},%Department and Organization
            city={Guangzhou},
            postcode={510006}, 
            country={China}}

\affiliation[inst3]{organization={Department of Computer Science, University of Surrey},%Department and Organization
            city={Guildford},
            postcode={GU2 7XH},
            country={United Kingdom}}

\begin{abstract}
\textcolor{black}{In recent years, there has been a notable surge in research on machine learning techniques for combinatorial optimization. It has been shown that learning-based methods outperform traditional heuristics and mathematical solvers on the Traveling Salesman Problem (TSP) in terms of both performance and computational efficiency. However, most learning-based TSP solvers are primarily designed for fixed-scale TSP instances, and also require a large number of training samples to achieve optimal performance. To fill this gap, this work proposes a data-driven graph representation learning method for solving TSPs with various numbers of cities. Specifically, we formulate the TSP as a link prediction task and propose an edge-aware graph autoencoder (EdgeGAE) model that can solve TSPs by learning from various-scale samples with an imbalanced distribution. A residual gated encoder is trained to learn latent edge embeddings, followed by an edge-centered decoder to output link predictions in an end-to-end manner. Furthermore, we introduce an active sampling strategy into the training process to improve the model's generalization capability in large-scale scenarios. To investigate the model's practical applicability, we generate a scale-imbalanced dataset comprising 50,000 TSP instances ranging from 50 to 500 cities. The experimental results demonstrate that the proposed edge-aware graph autoencoder model achieves a highly competitive performance among state-of-the-art graph learning-based approaches in solving TSPs with various scales, implying its remarkable potential in dealing with practical optimization challenges.}

\end{abstract}


% \begin{highlights}
% \end{highlights}


\end{frontmatter}



\section{Introduction}

Combinatorial optimization problems (COPs) exist in many practical applications of operational research (OR). Most COPs are NP-hard and difficult to solve within a polynomial time \citep{bengio2021machine}. As an important field of research on COPs, routing problems arise in various application scenarios such as logistics, transportation, urban planning, and supply chain management. Generally, routing problems involve planning route scheduling solutions for a group of vehicles under constrained conditions to make deliveries or provide services to a set of locations efficiently. The target is to minimize the total traveling distance or overhead with all constraints being satisfied \citep{xu2022reinforcement}. The traveling salesman problem (TSP) is one of the most intensively studied routing problems, which aims to find the shortest path to traverse a set of given locations once and return to the initial location. 

As an NP-hard problem, it is non-trivial to find an efficient polynomial-time solution via an enumeration search for TSPs with even only hundreds of cities. \textcolor{black}{Traditional methods for solving TSPs and other optimization problems can be generally categorized into mathematical solvers \citep{sanches2017improving,rosenkrantz1977analysis,helsgaun2017extension} and hand-crafted heuristics \citep{cui2024multi, wang2023multi, wang2023enhancing, wang2022solving}.} A representative of exact solvers is Concorde \citep{sanches2017improving}, which incorporates both cutting-plane and branch-and-bound approaches for solving TSPs. For small and medium-sized problems, exact TSP solvers can find optimal solutions with guarantees. However, for large-scale instances, it will be intractable for exact solvers to get an exact solution within a reasonable amount of time due to the exponential growth of the computational complexity. To alleviate this limitation, evolutionary algorithms (EAs) are adopted to solve TSPs and proven to be well-suited for exploring a large search space. However, population-based EAs often require a large number of function evaluations, leading to huge computational costs, especially when the evaluations are time-consuming. Additionally, EAs are unable to learn from existing historical data, and even minor modifications to the problem require restarting the search from scratch.

An alternative, known as neural combinatorial optimization (NCO) \citep{bello2017neural,bengio2021machine}, has been explored in the machine learning community from the perspective of data-driven optimization \citep{jin2018data,jin2021data,bengio2021machine}. Instead of hand-crafted heuristics, NCO trains neural networks in an end-to-end manner to solve any combinatorial problems by learning directly from existing instances. Early work on learning-based approaches for routing problems can be traced back to Hopfield networks \citep{hopfield1985neural} and template models \citep{angeniol1988self}. With the compelling performance of sequence-to-sequence models \citep{sutskever2014sequence,wu2016google}, Vinyals et al. proposed a Pointer Network \citep{vinyals2015pointer} with an attention mechanism \citep{bahdanau2014neural} to output TSP tours in a sequential manner. In a similar work proposed by Bello et al. \citep{bello2017neural}, the Pointer Network is trained by an Actor-Critic reinforcement learning strategy instead of supervised learning \citep{vinyals2015pointer}. 

With the development of graph representation learning \citep{chen2020graph, zhao2023obfuscating, zhao2023self}, NP-hard routing problems are reconsidered as sequential decision tasks on graphs \citep{joshi2019efficient}. Consequently, there has been a recent surge of work on training graph neural networks (GNNs) \citep{kipf2017semi,velivckovic2018graph,xu2018powerful} to solve COPs such as routing problems \citep{nowak2018revised,joshi2019efficient,joshi2022learning}, facility location problems \citep{liu2023end} and graph coloring problems \citep{wang2023graph}. Compared to OR solvers, GNN-based methods have achieved competitive performance on small problem instances \citep{joshi2022learning,joshi2019efficient,dai2017learning,li2018combinatorial}. Once trained, a GNN-based solver can generate approximate optimal solutions in a very short period of inference time, making it favorable for real-time decision-making. End-to-end learning for TSPs can be categorized into autoregressive and non-autoregressive approaches, depending on how solutions are generated. In autoregressive methods, a TSP tour is constructed regressively by identifying one city at a step \citep{dai2017learning,bello2017neural,deudon2018learning,kool2019attention}. By contrast, non-autoregressive approaches make predictions for arbitrarily large graphs in a one-shot fashion \citep{nowak2018revised,joshi2019efficient,dwivedi2020benchmarking,joshi2022learning}. In an early work \citep{nowak2018revised}, Nowak et al. trained GNNs to output the prediction as an adjacency matrix of the input graph, which is further converted into feasible solutions by beam search. Joshi et al. \citep{joshi2019efficient} built on ideas from \citep{nowak2018revised} and introduced the residual gated graph ConvNet \citep{bresson2017residual} for solving TSP. 
%The model is trained in a supervised manner with solution pairs from the Concorde solver and outputs an edge heatmap to identify the probability of each edge belonging to the optimal tour. The predictions are finally transformed into feasible solutions with the beam search strategy. 
By extending the work in \citep{joshi2019efficient}, Fu et al. \citep{fu2021generalize} replaced the beam search with reinforcement learning techniques. The output heatmaps are fed into the Monte Carlo tree search for TSP tours. 

However, existing work on data-driven NCO approaches for TSPs suffers from distinct drawbacks. First, most learning-based solvers are only trained on fixed-sized TSP instances \citep{kool2019attention,deudon2018learning,joshi2019efficient,vinyals2015pointer}, which exhibit poor generalization ability on test cases with a number of cities different from that of the training data \citep{joshi2022learning}. Secondly, training such a solver often requires millions of TSP instances with optimal solutions to achieve a competitive performance, assuming that the model has sufficient labeled data as its training dataset. However, it is non-trivial to satisfy this assumption in practice since NP-hard problems usually have a limited number of imbalanced optimized instances, which contain more small-scale cases and fewer large-scale cases.

In this paper, we propose an edge-aware graph autoencoder (EdgeGAE) framework for solving TSPs by learning from imbalanced data. \textcolor{black}{In contrast to the traditional TSP solvers that take TSP as a route planning task and decide the next city to visit step by step, this work reformulates TSP as a link prediction task on the graph. Each city is treated as a node on the graph, and the optimal route can be represented by the connections between nodes. The encoder-decoder structure is widely adopted for traditional link prediction tasks in graph representation learning. Nevertheless, the conventional graph autoencoder (GAE) fails to solve TSP tasks, as its encoding-decoding process only considers node features, overlooking the importance of edge information for TSP.} Considering that the quality of solutions to routing problems relies highly on the edge features (i.e., Euclidean distances), we modify the encoder with residual gated graph convolutions for the message-passing process to learn explicit edge embeddings in the latent space. To alleviate the limitation of inner products where only nodes are involved, we design an edge-centered decoder model that incorporates both the edge and corresponding node embeddings for the link prediction task. We adopt an active sampling (AS) strategy during the training process by incorporating \textit{oversampling} and \textit{undersampling} to deal with the imbalanced distribution of training data. The proposed framework is evaluated on TSP instances with different sizes up to hundreds of cities and compared with state-of-the-art learning-based methods in terms of the optimal gap and the F1 score. \textcolor{black}{Distinct from the traditional learning-based TSP solvers, }the main contributions of this work are summarized as follows:

\begin{enumerate}
  \item %We model the TSP as a link prediction task on graphs, and propose an edge-aware graph autoencoder model with enhanced graph representations and an edge-centered decoding scheme to solve TSPs by learning on data with an imbalanced distribution of various scales. 
  \textcolor{black}{We first transform the TSP from a traditional path planning problem to a link prediction task in the field of graph representation learning. Drawing on this, we redesign the structure of the vanilla graph autoencoder and propose an edge-aware graph autoencoder model to accommodate the TSP characteristics.}
  
  \item  %We perform random active sampling strategies on imbalanced data during the end-to-end training process to enhance the model's capability of generalizing to various scales. Comprehensive experiments demonstrate that the proposed model outperforms other learning-based methods on solving TSPs.
  \textcolor{black}{We tackle the challenge of extremely imbalanced data distribution by introducing a random active sampling strategy into the end-to-end training process. The incorporation of \textit{oversampling} and \textit{undersampling} enables our model to retain strong generalization performance with only a few large-scale training samples.}
  
  \item %We generate a benchmark dataset consisting of solution pairs of 50,000 TSP instances of different sizes ranging from 50 to 500 cities. The optimal solutions of the instances are generated by Concorde, and the data follows an extremely imbalanced distribution where the number of instances is inversely proportional to the number of cities, which is reasonable due to the limited computational resources.
 \textcolor{black}{To validate the model performance in scale-imbalanced optimization scenarios, we construct a benchmark dataset comprising 50,000 TSP instances with varying scales, ranging from 50 to 500 cities. The data follows an extremely imbalanced distribution where the number of instances is inversely proportional to the number of cities. Comprehensive experiments based on the benchmark demonstrate that the proposed EdgeGAE outperforms other learning-based methods for solving TSPs.}
  
\end{enumerate}

The remainder of this article is organized as follows. Section II gives a literature review of existing work on learning-based NCO methods for TSPs and their generalization ability. Section III provides the problem formulation and elaborates the proposed methodology in detail. The experimental results and analyses of these results are presented in Section IV. Section V concludes the paper and outlines future work.

\section{Related Work}

Several attempts have been made to improve the generalization ability of machine learning methods for routing problems such as TSPs \citep{fu2021generalize, li2023learning, luo2023neural, zhou2023towards}. An intuitive approach is to decompose a large-scale problem into a group of sub-problems and solve them separately. Fu et al. \citep{fu2021generalize} proposed to train a graph convolutional-based model on small-scale instances and generalize the pre-trained model to instances whose size is larger than that of the training data by using graph sampling and heatmap merging techniques. Specifically, a large number of instances with a fixed small size are first randomly generated and optimized by Concorde. Then a graph convolutional network \citep{bresson2017residual} is trained on the generated data in a supervised manner, following the same model and framework in \citep{joshi2019efficient}. The trained model outputs a heatmap for the input graph as the probability prediction of each edge appearing in the optimal solution. When testing on TSP instances with a larger scale than those seen in the training dataset, the input graph is first decomposed into smaller subgraphs which consist of exactly the same number of vertices as the training data. The pre-trained model is adopted to output heatmaps for each subgraph separately, and then these heatmaps are merged into a complete heatmap as the edge prediction for the original input instance. Finally, the Monte Carlo tree search \citep{browne2012survey} strategy is adopted to convert the heatmaps into feasible solutions to TSPs. \textcolor{black}{Similarly, Luo et al. \citep{luo2023neural} proposed a Light Encoder and Heavy Decoder (LEHD) model. The LEHD model adaptively captures the relationships among nodes of varying sizes dynamically, enhancing its ability to generalize across problems of diverse scales. }

While such decomposition-based approaches are simple to implement, they also come with several limitations. The pre-trained model can only be trained on graphs of a fixed size with a predefined number of vertices ($m$), which may introduce an additional process to identify the optimal hyperparameter $m$. When tackling TSP instances of an arbitrarily large size, it is required to decompose a large graph into subgraphs that have exactly the same number of $m$ vertices as the training data. As a result, it can easily lead to overlaps between different subgraphs, posing difficulty in solving the original problem. Additionally, the complete heatmap is created by combining a number of sub-heatmaps that only consider the corresponding local information, overlooking the global knowledge of the original graph. Finally, it should be noted that the decomposition-based approach requires a huge amount of training data, for example, 990,000 TSP instances are generated for the pre-trained model in \citep{fu2021generalize}. \textcolor{black}{An alternative is to transfer the knowledge learned from small-scale instances into the optimization process of large-scale instances. For example, Zhang et al. \citep{zhang2023neural} introduced progressive distillation into the training of NCO models by adopting curriculum learning to train TSP samples in an increasing order of the problem sizes.}

Other work combines deep learning methods with traditional heuristics to achieve a good generalization ability on large-scale problems \citep{bengio2021machine, xin2021neurolkh, zheng2021combining}. Traditional heuristic approaches are delicately designed with great interpretability based on decades of expert knowledge and experience \citep{colorni1996heuristics}. With the advances in machine learning techniques, powerful network models are trained to learn inherent knowledge and extract complex patterns from known instances to enhance the performance of traditional heuristics \citep{bengio2021machine}. \textcolor{black}{Ye et al. \citep{ye2023deepaco} proposed a neural-enhanced meta-heuristic algorithm named DeepACO to leverage reinforcement learning in the heuristic designs of ant colony optimization.} NeuroLKH is a representative work proposed by Xin et al. \citep{xin2021neurolkh}. In NeuroLKH, a sparse graph network (SGN) is combined with the strong heuristic Lin-Kernighan-Helsgaun (LKH) \citep{helsgaun2000effective,helsgaun2009general} for solving TSP instances. In the traditional LKH, an edge candidate set is pre-defined based on hand-crafted rules for the $\lambda$-opt searching process \citep{lin1965computer}, and node penalty values are iteratively optimized via sub-gradient optimization to transform the edge distances for the searching process. NeuroLKH trains an SGN model to generate edge scores and node penalties simultaneously. Then the edge scores are used to create an edge candidate set instead of hand-crafted rules, and the node penalties are used to transform edge distances without performing iterative optimization for each instance as the traditional LKH does. The SGN model is trained on a wide range of instances with different sizes, which are optimized by the exact solver Concorde. When generalized to larger instances, the edge scores in SGN can be directly adopted without any modification, but a fine-tuning step is required to adapt the scale of the node penalties to large sizes. Similarly, VSR-LKH \citep{zheng2021combining} combines three reinforcement learning strategies (Q-learning \citep{watkins1992q}, Sarsa \citep{sutton2018reinforcement} and Monte Carlo \citep{browne2012survey}) with the traditional LKH algorithm for solving TSP. It replaces the fixed traversal operation in LKH and enables the algorithm to choose each search step by learning from reinforcement learning strategies. \textcolor{black}{Inspired by the great success achieved by Large Language Model (LLM) \citep{wu2024evolutionary} recently, Liu et al. \citep{liu2024example} proposed an evolutionary optimization algorithm assisted by a Large Language Model by incorporating the LLM into the paradigm of evolutionary computation. The proposed method can automatically evolve the elite-guided local search algorithms which outperform human-designed algorithms in solving small-scale TSPs. Wang et al. \citep{wang2024asp} proposed a novel approach named Adaptive Staircase Policy Space Response Oracle to address the generalization issues of NCO, aiming to help neural solvers explore and adapt to different distributions and various problem scales.}

A recent research in \citep{joshi2022learning} brings together several recent works \citep{kool2019attention,joshi2019efficient,nowak2018revised,deudon2018learning} on learning-based TSP approaches into a uniform pipeline, with the aim to investigate various components in the algorithm that influence the generalization ability, including the model construction, learning paradigms and inductive biases \citep{battaglia2018relational}. To this end, the TSP is considered as a sequential decision-making task on graphs \citep{joshi2019efficient}, and a unified end-to-end learning pipeline is constructed with five stages. The first stage is the problem definition, where the TSP is formulated as a fully connected graph \citep{dai2017learning}. The weight value of each edge can be directly determined by the Euclidean distance according to the coordinates of the cities. For better computational efficiency and generalization, the graph can be sparsified via a priori knowledge such as $k$-nearest neighbours \citep{dwivedi2020benchmarking,joshi2019efficient,fu2021generalize}. The second stage is graph embedding, where the GNN models are usually trained to output latent space embeddings for nodes and edges of the input graph \citep{kipf2017semi,gilmer2017neural,velivckovic2018graph}. Solution decoding and search process takes place in the next two stages. The latent embeddings are transformed into probability values for each edge belonging to the optimal solution set via either autoregressive \citep{dai2017learning,kool2019attention,deudon2018learning,ma2019combinatorial,kwon2020pomo} or non-autoregressive manners\citep{joshi2019efficient,nowak2018revised,fu2021generalize,kool2022deep,dwivedi2020benchmarking}. Then feasible solutions to the original problem are generated via graph search approaches such as beam search and greedy search \citep{joshi2019efficient,freitag2017beam,huber2022learning}. Finally, the entire model is trained in an end-to-end manner by imitating an expert via supervised learning \citep{dwivedi2020benchmarking, joshi2019efficient}, or by minimizing a cost function via reinforcement learning \citep{kool2019attention,bello2017neural}. The controlled experiments in \citep{joshi2022learning} reveal several insights into zero-shot generalization. Concretely, the dominant evaluation approach tends to obscure the limited generalization ability, as the model performance is evaluated on fixed or excessively small TSP instances. The generalization performance of the GNN-based models benefits from dedicated redesign by considering shifting graph distributions. Compared to non-autoregressive decoding, the autoregressive manner improves generalization by imposing a sequential bias. However, it significantly increases the inference time, especially as the size of the instances grows.

\section{The Proposed Methodology}
In this section, we propose our learning-based EdgeGAE approach for solving TSPs of various sizes from a link prediction perspective. We first define the TSP on graphs and outline the overall framework. Then we develop a novel edge-aware residual graph autoencoder model with enhanced graph representation and edge-centered decoding scheme. Considering the various scales and imbalanced distribution of the training data, we design a training strategy with batch encoding and active sampling to improve the generalization ability. Finally, we construct a TSP benchmark dataset composed of 50,000 instances-solution pairs generated by Concorde to simulate the scale-imbalanced distribution in practical scenarios.

\begin{figure*}
    \centering
    \setlength{\abovecaptionskip}{0.cm}
    \includegraphics[width=1.0\textwidth]{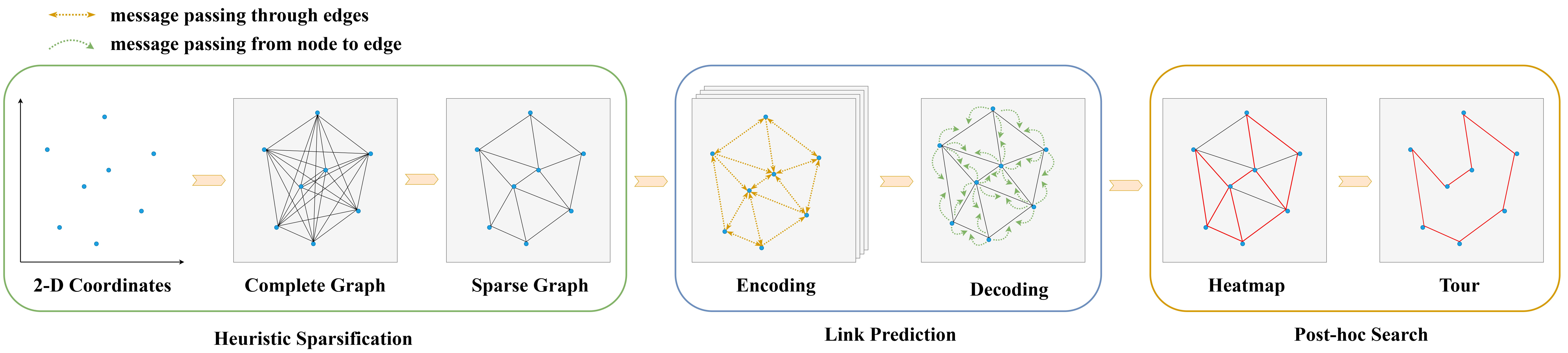}
    \caption{Solving TSPs as a link prediction task on graphs. In the heuristic sparsification stage, the input coordinates of $N$ cities are first converted into a complete graph where all nodes are directly connected, followed by heuristics such as $k$-nearest neighbours to get a sparse graph without redundant edges. In the link prediction stage, the sparse graph with node and edge features is input to the encoder to get latent graph embeddings via multiple layers of message passing through edges (in yellow arrows). Then the decoder performs link prediction on each edge by aggregating the embeddings of the source and target node into the edge embedding via an edge-centered message passing scheme (in green arrows). Finally, the output heatmaps are converted into feasible solutions through post-hoc graph search approaches.}
    \label{fig:framework}
\end{figure*}

\subsection{Overall Framework}  % Problem Formulation
\label{sec:overall_framework}

In this paper, we consider the 2-D Euclidean symmetric TSP, where each city has a two-dimensional coordinate, and the distance between two cities is independent of the traveling direction. It aims at finding the shortest route for a salesman to visit each city once and return to the departure city. Given a TSP instance with $N$ cities, the input is a set of order-invariant nodes $S=\left\{x_i\right\}_{i=1}^N$ where $x_i \in \mathbb{R}^2$ represents the 2-D coordinates of the $i$-th node. A feasible solution to the problem is a permutation of $N$ nodes $\bm{\pi}=\left\{\pi_1, \pi_2, ..., \pi_N \right\}$ where $\pi_i$ is the index of the $i$-th node in the permutation. Assume that the salesman starts from city $x_{\pi_1}$ and traverses all the cities in the order of the permutation $\bm{\pi}$ until $x_{\pi_N}$ and finally returns to the departure city. The objective is to minimize the total length $L(\bm{\pi} \mid S)$ of the tour:

\begin{equation}
\label{eq1}
    L(\bm{\pi} \mid S)=\left\|x_{\pi_N}-x_{\pi_1}\right\|_2+\sum_{i=1}^{N-1}\left\|x_{\pi_i}-x_{\pi_{i+1}}\right\|_2,
\end{equation}
where $x_{\pi_i}$ denotes the 2-D coordinates of the $i$-th visited city and $\|\cdot\|_2$ symbolizes the $\ell_2$ norm. 

In contrast to the autoregressive methods that consider TSPs as a sequential decision-making process, we formulate it as a binary-class link prediction task on graphs. The overall framework of the proposed methodology is illustrated in Figure \ref{fig:framework}. 

In earlier work \citep{kool2019attention,joshi2019efficient}, TSP instances are represented as complete graphs where there are connections between each pair of nodes (cities) no matter what the Euclidean distance is. In a complete graph representation $\mathcal{G}(\mathcal{V, E})$, $\mathcal{V}$ denotes the set of $N$ cities as nodes with $|\mathcal{V}|=N$, and $\mathcal{E}$ consists of all undirected edges connecting any two nodes in the set $\mathcal{V}$. The number of connections $|\mathcal{E}|$ increases exponentially with the number of cities, resulting in a high density of edges. Consequently, the pairwise computation for all nodes will be intractable as the graph becomes larger. Furthermore, according to the definition of the TSP, the connections in an optimal solution are more likely to exist between two nodes that are close to each other, as they usually result in a smaller objective value.

To improve the scalability of the model while preserving valuable information on the graph topology, we adopt $k$-nearest neighbours heuristics and convert the TSP instance from a complete graph to a sparse graph representation $\hat{\mathcal{G}}(\mathcal{V},\hat{\mathcal{E}})$ with each node only connected to its $k$ nearest neighbors:

\begin{equation}
\label{eq2}
    \hat{\mathcal{E}} = \{(u, v) \,|\, u, v \in \mathcal{V}, v \in U_{k}\},
\end{equation}
where $U_{k}$ is the set of all $k$-nearest neighbours of node $u$ according to the Euclidean distances. The heuristic graph specification enables learning faster and scaling up to large instances by reducing the search space. In $\hat{\mathcal{G}}(\mathcal{V},\hat{\mathcal{E}})$, the node features are defined as the normalized 2-D coordinates, and the edge features are the corresponding Euclidean distance between the source and target nodes. We transform the optimization of a TSP instance into a binary-class link prediction task and define the ground-truth label $Y(\mathbf{e})$ of each edge $\mathbf{e}$ as whether it exists or not in the optimal tour $\bm{{\pi}^*}$.

\begin{equation}
    Y(\mathbf{e} \mid \mathbf{e} \in \hat{\mathcal{E}})= \begin{cases}1, & \text { edge } \mathbf{e} \text { exists in } \bm{{\pi}^*}; \\ 0, & \text { otherwise. }\end{cases}
\end{equation}

In data-driven NCO, a neural network will be trained with labeled data (instances) in an end-to-end manner. After training, the model can output link predictions $\hat{Y}(\mathbf{e})$ on all edges of the input graph.

\begin{equation}
    \hat{Y}(\mathbf{e} \mid \mathbf{e} \in \hat{\mathcal{E}}) \in [0,1]
\end{equation}

Finally, the output predictions are transformed into feasible solutions by graph search approaches. In this work, we employ the basic random search and 2-opt as our post-hoc search strategy \citep{croes1958method}. Given the link predictions in the form of a heatmap, we construct a TSP tour sequentially by adding one node at each step. Starting from a randomly selected node, at each step we calculate the normalized prediction values of all unvisited nodes as the probability of being selected in the next step. Then the next node is selected via roulette and added to the partial tour. The selection continues recursively until all nodes are visited, followed by a 2-opt local search to further improve the quality of the solution. The 2-opt evaluates all possible edge swaps at each iteration, and selects the swap that can lead to the greatest reduction in the total length of the tour until no further improvements can be made. The post-hoc search algorithm with random sampling and 2-opt local search is shown in Algorithm \ref{alg:posthoc}.

\begin{algorithm}[htbp]
\caption{Post-hoc Graph Search with 2-opt Heuristics}
\label{alg:posthoc}
\begin{algorithmic}[1]
\State \textbf{Input}: Link predictions as a heatmap
\State \textbf{Output}: TSP tour
\State Initialize an empty tour $tour$
\State Randomly select a start node $n_0$ from $N$ nodes
\State Add $n_0$ to $tour$
\State Mark $n_0$ as visited

\For{$i \gets 1$ to $N-1$}
\State Normalize prediction values for all unvisited nodes
\State Identify the next node $n_i$ using roulette selection
\State Add $n_i$ to $tour$
\State Mark $n_i$ as visited
\EndFor
% ----------------------------------------
\State $improved \gets True$
\While{$improved$ is $True$}
\State $improved \gets False$
\For{$i \gets 1$ to $N-2$}
\For{$j \gets i+1$ to $N-1$}
\State $newTour \gets$ reverse the order of nodes between $n_i$ and $n_j$ in $tour$
\If{length($newTour$) $<$ length($tour$)}
\State $tour \gets newTour$
\State $improved \gets True$
\EndIf
\EndFor
\EndFor
\EndWhile
\State \textbf{return} $tour$
\end{algorithmic}
\end{algorithm}

\subsection{The Edge-aware Residual Graph Autoencoder}

\begin{figure*}
    \centering
    \includegraphics[width=1.0\textwidth]{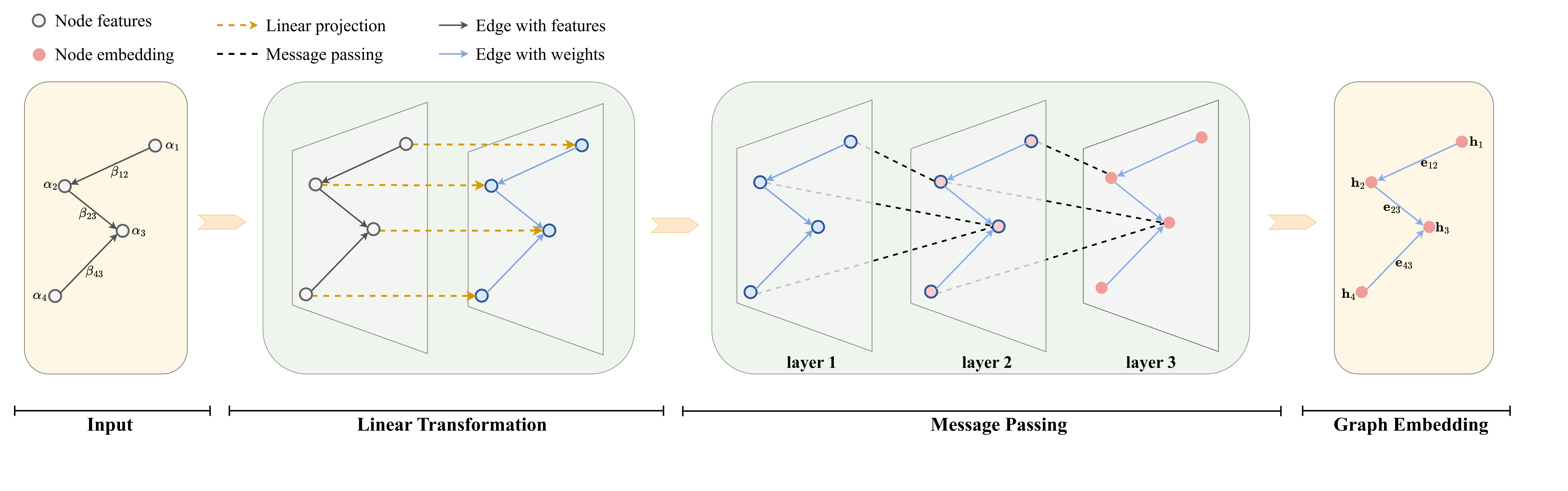}
    \caption{The residual gated encoder. (1) The input to an encoder model is a directed sparse graph, where the input node features and edge features are represented as $\alpha$ and $\beta$, respectively. In the sparse graph representation of the TSP, each node is connected to its $k$ nearest neighbours. (2) The node and edge features are mapped to high-dimensional embeddings through linear projections to enhance the expression power. (3) Multiple message-passing layers enable nodes and edges to learn information from local neighbourhoods via residual gated graph convolutions. (4) The encoder outputs the graph with latent space embeddings which can capture inherent knowledge of the graph topology for downstream tasks.}
    \label{fig:encoder}
\end{figure*}

\subsubsection{A Residual Gated Encoder}
The encoder module focuses on learning enhanced latent space graph embeddings via an efficient message-passing process. The message passing is a fundamental operator on graph-structured data in the encoding scheme, which enables effective information exchange and aggregation among nodes based on the topology of the graph. To capture complex graph patterns via information propagation, the message passing involves a recursive process of broadcasting and updating messages between related nodes in the graph. And the recursive message passing which can map an input graph to its hidden space representation is usually achieved by multiple layers of graph convolution-based operations. Specifically, at each layer, a node aggregates information from its neighbour nodes which have direct connections (edges) with itself, and then calculates and updates its representation according to the integrated information. Such an aggregation and update process occurs at all nodes simultaneously at each step, and this process is repeated for multiple iterations on the graph before generating the final representation. The motivation of a recursive message-passing process is to enable every node to refine its embeddings for capturing progressively implicit patterns. Within one layer of the message passing, a node learns information from its $1$-hop neighbours. After $k$ layers of propagation, each node gets a reception field of its $k$-hop neighbour nodes. Therefore, the recursive message passing encourages the node embeddings to capture high-level dependencies by aggregating messages from distant nodes, and promotes the encoder to learn hierarchical representations for downstream tasks. Consequently, the depth of the encoder, measured by the number of layers, is crucial for effective feature extraction and message propagation.

\textcolor{black}{As illustrated in Figure \ref{fig:encoder},} there are mainly three steps in the message passing process of the encoder: $transformation$, $aggregation$, and $update$. In the message transformation step, all neighbouring nodes encode their features as messages, which may contain node-specific information or relationships. Then the embeddings are mapped to latent space representations via a linear or non-linear transformation. The mapped embeddings can have either the same or different dimensions as before. In the message aggregation step, each node aggregates the transformed information from its $1$-hop neighbours according to the graph topology. Some common choices for aggregation functions include $sum$, $mean$, and $max$. More complex alternatives include weighted aggregation or attention mechanisms. The aggregation function should be permutation invariant to ensure that the aggregation result remains the same regardless of the processing order of nodes. Once the messages are aggregated, in the message update step, each node calculates a refined version of its representation via an update function that takes both the original embedding and the aggregated information as the inputs. Finally, the updated representations are taken as the inputs to the next layer of the encoder. The message-passing process can be generally formulated as:
\begin{equation}
    \label{eq:mp}
    \mathbf{h}_i \leftarrow \gamma\left(\mathbf{h}_i, \bigotimes_{j \in \mathcal{N}_i} \phi \left(\mathbf{h}_i, \mathbf{h}_j, \mathbf{e}_{j, i}\right)\right),
\end{equation}
where $\mathbf{h}_i$ denotes the embedding of the $i$-th node in the graph and $\mathbf{e}_{j, i}$ denotes the embedding of a directed edge from node $j$ to node $i$. $\phi$ is the message transformation function that takes the original node or edge embeddings as the inputs, and $\bigotimes$ represents a permutation invariant function such as $sum$. $\mathcal{N}_i$ consists of all $1$-hop neighbours of node $i$, and $\gamma$ is the update function which calculates the updated embedding $\mathbf{h}_i$ according to the original embedding and the aggregated message from its first-order neighbouring nodes.

As formulated in Equation \ref{eq:mp}, the message exchange occurs at the local neighbours of each node in the graph, enabling them to receive inherent information about the structure and node features across the entire graph. The expressive capability of latent embeddings plays an important role in the following link prediction task, which requires a good understanding of the relationships and interconnections in the graph. However, the original graph autoencoder framework \citep{kipf2016variational} only takes a simple graph convolutional network (GCN) as its encoder model. The layer-wise propagation rule is as follows:
\begin{equation}
    \label{eq:gcn}
    H^{(l+1)}=\sigma\left(\tilde{D}^{-\frac{1}{2}} \tilde{A} \tilde{D}^{-\frac{1}{2}} H^{(l)} W^{(l)}\right)
\end{equation}
\begin{equation}
    \tilde{D}_{i i}=\sum_j \tilde{A}_{i j},
\end{equation}
where $H^{(l)}$ denotes the node embeddings of the $l$-th GCN layer. $\tilde{A}$ and $\tilde{D}$ are the adjacency matrix with self-connections and the degree matrix, respectively. $\sigma$ represents the activation function and $W^{(l)}$ is the learnable weight matrix of the $l$-th layer. 
In the GCN-based encoder model, only node embeddings are propagated and updated within message-passing layers, without accounting for any edge features. In addition, the same message transformation matrix $W^{(l)}$ is shared among all nodes, limiting the learning capability of the encoder model to identify fine-grained relationships between nodes. Furthermore, the iterative propagation in the standard GCN could result in over-smoothing, significantly reducing the diversity among different nodes after several message-passing layers. To overcome these limitations, we design an edge-aware encoder model with residual connections and a dense-attention mechanism built on ideas from graph ConvNets \citep{bresson2017residual}. 

For a TSP instance with $N$ cities, its graph representation $\mathcal{G}(\mathcal{V},\mathcal{E})$ is taken as the input of the encoder. We first map the original node and edge features to high-dimensional embeddings via linear transformations:

\begin{equation}
    \label{eq:node_emb}
    \mathbf{h}_i=\mathbf{W}_h \cdot \bm{\alpha}_i+\mathbf{b}_h, \quad  \bm{\alpha}_i \in \mathcal{V}
\end{equation}

\begin{equation}
    \label{eq:edge_emb}
    \mathbf{e}_{i,j}=\mathbf{W}_e \cdot \bm{\beta}_{ij}+\mathbf{b}_e, \quad \bm{\beta}_{ij} \in \mathcal{E},
\end{equation}
where $\bm{\alpha}_i \in \mathbb{R}^2$ is the input node feature of the $i$-th city denoting the 2-dimensional coordinates, and $\bm{\beta}_{ij} \in \mathbb{R}$ represents the Euclidean distance of the edge between cities $i$ and $j$. $\mathbf{W}_h \in \mathbb{R}^{H \times 2}$, $\mathbf{W}_e \in \mathbb{R}^{H \times 1}$, $\mathbf{b}_h \in \mathbb{R}^{H}$ and $\mathbf{b}_e \in \mathbb{R}^{H}$ are the trainable weight parameters of the linear transformations. $\mathbf{h}_i$ and $\mathbf{e}_{ij}$ are $H$-dimensional node and edge embeddings, respectively. Since the performance of link prediction for the TSP relies heavily on the total length of selected edges, we construct explicit edge embeddings via an edge convolutional operation in each layer of the encoder:
\begin{equation}
    \label{eq:edge_conv}
    \mathbf{e}_{ij}^{(k+1)} = \mathbf{C}^{k} \mathbf{e}_{ij}^{(k)} + \mathbf{D}^{k} \mathbf{h}_i^{(k)} + \mathbf{E}^{k} \mathbf{h}_j^{(k)},
\end{equation}
where $\mathbf{C}^{k}, \mathbf{D}^{k}, \mathbf{E}^{k} \in \mathbb{R}^{H \times H}$ are learnable parameters of the $k$-th edge convolution layer, and the index $k$ indicates that the weight parameters are not shared among different layers. The enhanced edge embeddings are further adopted to generate a set of dense attention-based weight vectors for node convolution.

\begin{equation}
    \label{eq:weight}
    \boldsymbol{\omega}_{i j}^k=\frac{\sigma\left(\mathbf{e}_{ij}^{(k)}\right)}{\sum_{j \in \mathcal{N}_i} \sigma\left(\mathbf{e}_{ij}^{(k)}\right)+\delta},
\end{equation}
where $\sigma(\cdot)$ denotes the sigmoid function and $\delta \in \mathbb{R}^{+}$ is a small constant. The node convolutional operator in the $k$-th encoder layer is written by:
\begin{equation}
    \label{eq:node_conv}
    \mathbf{h}_{i}^{(k+1)}=\mathbf{A}^{k} \mathbf{h}_i^{(k)}+\sum_{j \in \mathcal{N}_i} \boldsymbol{\omega}_{i j}^{k} \odot \mathbf{B}^{k} \mathbf{h}_j^{(k)},
\end{equation}
where $\mathbf{A}^{k}, \mathbf{B}^{k} \in \mathbb{R}^{H \times H}$ are trainable parameters of the node convolution, and $\boldsymbol{\omega}_{i j}^{k}$ is the gated dense-attention map vector derived from the edge embedding $\mathbf{e}_{ij}$. A batch normalization (BN) operator is incorporated into the message-passing process for both node and edge embeddings before the activation function ReLU. Finally, we add a residual connection within each encoder layer to alleviate the over-smoothing issue after multiple iterations of message passing. The overall message-passing process within each layer of our edge-aware encoder model is formulated as:
\begin{equation}
    \label{eq:resgated_node}
    \mathbf{h}_{i}^{(k+1)}=\text{ReLU}(\text{BN}(\mathbf{A}^{k} \mathbf{h}_i^{(k)}+\sum_{j \in \mathcal{N}_i} \boldsymbol{\omega}_{i j}^{k} \odot \mathbf{B}^{k} \mathbf{h}_j^{(k)})) + \mathbf{h}_{i}^{(k)}.
\end{equation}

\begin{equation}
    \label{eq:resgated_edge}
    \mathbf{e}_{ij}^{(k+1)}=\text{ReLU}(\text{BN}(\mathbf{C}^{k} \mathbf{e}_{ij}^{(k)} + \mathbf{D}^{k} \mathbf{h}_i^{(k)} + \mathbf{E}^{k} \mathbf{h}_j^{(k)})) + \mathbf{e}_{ij}^{(k)}.
\end{equation}

In contrast to the original GCN-based encoder, the edge-aware encoder model can generate explicit edge representations via multiple layers of an efficient message-passing process. This enables the model to learn distinctive edge embeddings directly from input features (distances). Since the objective function of the TSP is determined by the sum of distances in a tour, such an edge-aware encoding scheme is more effective for link predictions than the simple concatenation of source and target node representations. The graph encoding scheme for enhanced edge representations is illustrated in Algorithm \ref{alg:encoding}.

\begin{algorithm}
\caption{Graph Encoding Scheme}
\label{alg:encoding}
\begin{algorithmic}[1]
\State \textbf{Input:} Graph $\mathcal{G}(\mathcal{V},\mathcal{E})$ with node features $\bm{\alpha}_i \in \mathcal{V}$ and edge features $\bm{\beta}_{ij} \in \mathcal{E}$, number of message passing layers $K$, latent space dimension $H$
\State \textbf{Output:} latent node encodings $\mathbf{H} \in \mathbb{R}^{|\mathcal{V}| \times H}$ and latent edge encodings $\mathbf{E} \in \mathbb{R}^{|\mathcal{E}| \times H}$
\State Map node features to latent space: $\mathbf{H} \gets \bm{\alpha}$ in Equation.\ref{eq:node_emb}
\State Map edge features to latent space: $\mathbf{E} \gets \bm{\beta}$ in Equation.\ref{eq:edge_emb}
\For{$k=1$ to $K$}
    \For{$i=1$ to $|\mathcal{V}|$ \textbf{in parallel}}
    \State Edge convolution on $\left\{e_{i j}^{k} \mid j \in \mathcal{N}_i\right\}$ in Equation.\ref{eq:edge_conv}
    \State Calculate gated weight vectors $\boldsymbol{\omega}_{i j}^k$ in Equation.\ref{eq:weight}
    \State Calculate node convolution on $\mathbf{h}_{i}^{k}$ in Equation.\ref{eq:node_conv}
    \State Batch normalization on both edges and nodes:
    \State $\mathbf{h}_i^{(k+1)} \gets \text{BN}(\mathbf{h}_i^{(k+1)})$
    \State $\mathbf{e}_i^{(k+1)} \gets \text{BN}(\mathbf{e}_i^{(k+1)})$
    \State Add residual connections after activation function:
    \State $\mathbf{h}_{i}^{(k+1)} \gets \text{ReLU}(\mathbf{h}_i^{(k+1)}) + \mathbf{h}_{i}^{(k)}$
    \State $\mathbf{e}_{i}^{(k+1)} \gets \text{ReLU}(\mathbf{e}_i^{(k+1)}) + \mathbf{e}_{i}^{(k)}$
    \EndFor
\EndFor
\State \textbf{return} $\mathbf{H} \in \mathbb{R}^{|\mathcal{V}| \times H}$ and $\mathbf{E} \in \mathbb{R}^{|\mathcal{E}| \times H}$
\end{algorithmic}
\end{algorithm}

\subsubsection{An Edge-centered Decoder}

\begin{figure*}
    \centering
    \setlength{\abovecaptionskip}{0.cm}
    \includegraphics[width=1.0\textwidth]{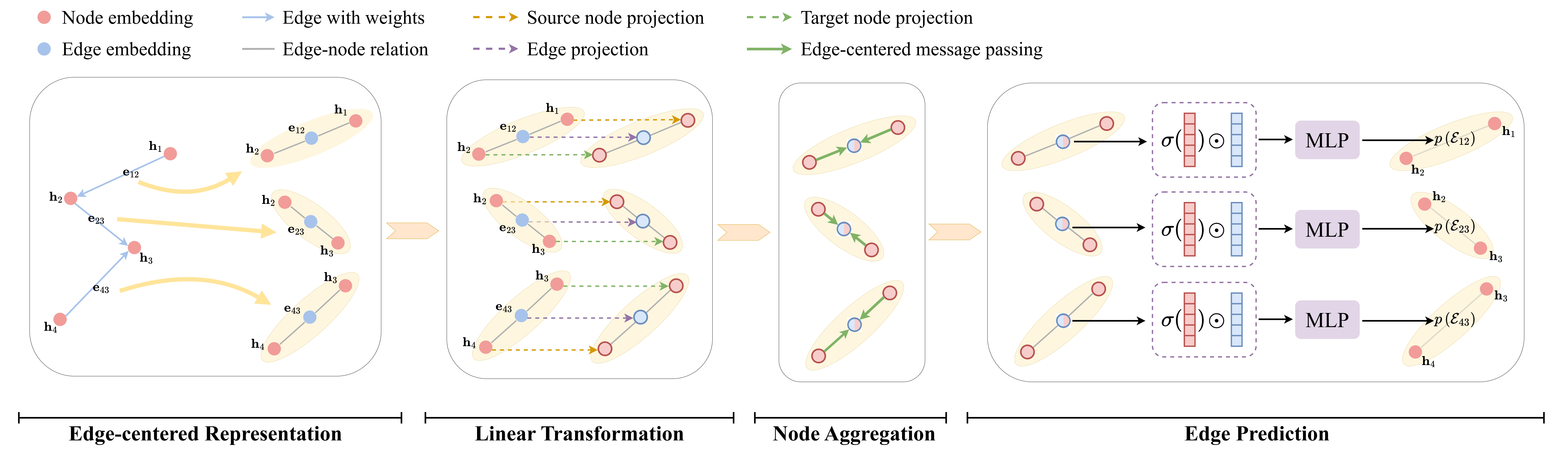}
    \caption{The edge-centered decoder. (1) We formulate the input graph of the decoder into an edge-centered representation for a clear illustration, where each edge embedding is associated with its source and target node embeddings. (2) Three different linear projections are applied to all source nodes, target nodes, and edge embeddings, respectively. (3) Embeddings from the source and target nodes of each edge are summed up, and then aggregated into the edge embedding in an edge-centered message-passing manner. (4) The inner product result of the sigmoid-transformed node embedding and the edge embedding is passed through a multi-layer perceptron (MLP) classifier to output the probability of the edge existing in the optimal tour.}
    \label{fig:decoder}
\end{figure*}

After encoding the node and edge representation in a latent space via the encoder model, the decoder aims to identify whether each edge exists in the optimal solution by making link predictions on the graph based on the encoding information. Generally speaking, a graph autoencoder model performs link prediction by leveraging the learned node embeddings to estimate the likelihood of a connection between two nodes of the input graph. In the traditional link prediction tasks such as Cora \citep{kipf2016variational}, Citeseer \citep{velivckovic2018graph} and PubMed \citep{sen2008collective}, there is only one graph in the dataset with thousands of nodes. The encoder-decoder model is trained on an incomplete version of the original graph as training data, where part of the edges are removed from the graph with all node features remaining. The encoder learns to map the input node features into latent embeddings, and the decoder reconstructs the graph based on the learned embeddings. The training target is to minimize the reconstruction error of the input subgraph. Once trained, the node embeddings are extracted from the encoder model for link prediction, which is supposed to have captured the node attributes and relationships of the entire graph. With the extracted embeddings, the decoder model estimates the likelihood of edge existence by calculating the similarity between two nodes. The original graph autoencoder model employs a simple inner product decoder to measure the similarity between a pair of nodes \citep{kipf2016variational}:
\begin{equation}
    \label{eq:innerproduct}
    p\left(\mathcal{E}_{ij} = 1 \mid \mathbf{h}_i, \mathbf{h}_j\right) = \sigma(\mathbf{h}_i^\top \mathbf{h}_j),
\end{equation}
where $p\left(\mathcal{E}_{ij}\right) \in [0,1]$ denotes the probability of an edge $\mathcal{E}_{ij}$ existing between node $i$ and node $j$. $\mathbf{h}_i, \mathbf{h}_j \in \mathbb{R}^H$ represent the learned node embeddings from the encoder model, and $\sigma(\cdot)$ is the sigmoid function. The value of the inner product between two node vectors implies their similarity in the latent space. According to Equation.\ref{eq:innerproduct}, the likelihood of the existence of an edge is determined by the degree of similarity between two node embeddings. A greater similarity between two nodes leads to a higher probability of a connection, and vice versa. This kind of similarity-based link prediction principle is based on the assumption that those nodes connected in a graph should have similar features and local neighborhood structures. This assumption applies to node-centered link prediction tasks like Cora, where similar documents do have a greater probability of cross-citation. However, it is not directly applicable to the optimal-tour prediction on graph-based routing problems like TSPs. 

Based on the problem definition in Section \ref{sec:overall_framework}, every node on a sparse TSP graph has the same degree determined by a heuristic value $k$, and the node features involve its two-dimensional coordinates in the Euclidean space. Therefore, all nodes on the graph have a similar local neighbourhood structure independent of the node characteristics. Furthermore, there is no guarantee that an edge between two nodes with similar features (i.e., a similar horizontal or vertical coordinate) must exist in the optimal tour. The objective value of a TSP solution is determined by the distances of all edges in the tour, without direct relations to the coordinates of cities. In summary, the original GAE with an inner product decoder performs well on node-centered link prediction tasks, since it encourages similar nodes to have similar embeddings, which is suitable for the graph characteristics in transductive datasets. On the other hand, it overemphasizes the node similarity and proximity at the expense of edge information, leading to a poor generalization ability to link prediction on TSP, which requires the incorporation of both edge and node messages.

For the above reasons, we redesign the decoder structure and propose an edge-aware decoder integrating both node and edge information for efficient link predictions on TSPs. \textcolor{black}{The decoder structure is shown in Figure \ref{fig:decoder}.} The general idea is to decode an edge representation by incorporating its source and target node embeddings via an edge-centered message-passing process. In contrast to the encoding process where messages are passed from one node to another along the edges, in edge-centered decoding, messages on source and target nodes are aggregated into the edge embedding to update itself. To this end, the following three main steps are implemented in the decoding scheme:
\begin{enumerate}[label=\alph*)]
    \item Message transformation. To enhance the expressive power of learned graph representations, a shared linear transformation is performed on all source nodes, target nodes, and edges separately.
    \item Node aggregation. For each edge in the graph, the transformed information on its source and target nodes are aggregated.
    \item Edge update. The aggregated node information is first normalized by the sigmoid function. Then, an inner product is performed on the node and edge embeddings.
\end{enumerate}

Following the message passing scheme in the proposed edge-aware decoder, the decoding output $\mathbf{d}_{ij}$ for an input edge embedding $\mathbf{e}_{ij}$ is calculated by:
\begin{equation}
    \mathbf{d}_{ij} = \sigma \left( \mathbf{F} \mathbf{h}_i + \mathbf{G} \mathbf{h}_j \right) \odot \mathbf{J} \mathbf{e}_{ij}
\end{equation}
where $\mathbf{F}, \mathbf{G}, \mathbf{J} \in \mathbb{R}^{H \times H}$ denote learnable parameters of the linear transformation for target nodes, source nodes, and edges, respectively, and $\sigma \left( \cdot \right)$ is the sigmoid function. Finally, the $H$-dimensional decoding is converted into a probability prediction via a simple MLP model. The edge-centered decoding scheme for predicting the TSP edges is described in Algorithm \ref{alg:decoding}.

\begin{equation}
    \label{eq:mlp}
    p\left(\mathcal{E}_{ij} = 1 \right) = \text{MLP} \left( \mathbf{d}_{ij} \right)
\end{equation}

\begin{algorithm}
\caption{Edge-centered Decoding Scheme}
\label{alg:decoding}
\begin{algorithmic}[1]
\State \textbf{Input:} Graph $\mathcal{G}(\mathcal{V},\mathcal{E})$, node embeddings $\mathbf{H} \in \mathbb{R}^{|\mathcal{V}| \times H}$, edge embeddings $\mathbf{E} \in \mathbb{R}^{|\mathcal{E}| \times H}$
\State \textbf{Output:} link prediction results
\For{$i=1$ to $|\mathcal{V}|$ \textbf{in parallel}}
    \For{$j \in \mathcal{N}_i$ \textbf{in parallel}}
        \State Transform messages on nodes and edges:
        \State $\hat{\mathbf{h}}_i \gets \mathbf{F}\mathbf{h}_i$, $\hat{\mathbf{h}}_j \gets \mathbf{G}\mathbf{h}_j$, $\hat{\mathbf{e}}_{ij} \gets \mathbf{J}\mathbf{e}_{ij}$
        \State Aggregation from source and target nodes:
        \State $ \hat{\mathbf{h}}_{agg} = \hat{\mathbf{h}}_i + \hat{\mathbf{h}}_j$
        \State Update the edge embedding:
        \State $\mathbf{d}_{ij} = \sigma ( \hat{\mathbf{h}}_{agg} ) \odot \mathbf{J} \mathbf{e}_{ij}$
        \State Calculate the existence probability:
        \State $p\left(\mathcal{E}_{ij} = 1 \right) = \text{MLP} \left( \mathbf{d}_{ij} \right)$
    \EndFor
\EndFor
\State \textbf{return} $ \left\{p\left(\mathcal{E}_{ij} = 1 \right) \mid \mathcal{E}_{ij} \in \mathcal{E}  \right\}$
\end{algorithmic}
\end{algorithm}

\subsection{Training Strategy}
\label{sec:active_sampling}
In data-driven NCO approaches, it is always challenging to handle data with various scales and imbalanced distributions. A typical scenario is combinatorial optimization in real-world applications, where the number of historical cases of small scales is much larger than the cases of large scales. Learning algorithms often face difficulties in generalizing inductive rules due to this kind of imbalance. Therefore, dealing with such problems requires the data-driven model to have the ability to learn from scale-imbalanced data and generalize to instances of arbitrary scales. In general, a model trained on small-scale instances tends to have a significant degradation in performance on large-scale problems, since the implicit knowledge learned from small-scale data cannot be directly transformed to large-scale problems. It has been demonstrated that learning from data with variable sizes is beneficial for the model's generalization ability \citep{joshi2022learning}. However, training a generalizable NCO model with limited large-scale instances while effectively leveraging knowledge from small-scale data remains an open question. 

In this work, we introduce a simple yet effective training strategy called \textit{random active sampling} \citep{he2009learning} into our training process, in order to learn from scale-imbalance TSP instances efficiently for link prediction tasks. The active sampling methods are originally designed to address imbalanced classification problems, where a majority class $D_{maj}$ has significantly more data than other minority classes $D_{min}$. Since an equalized dataset generally improves the overall classification accuracy than an imbalanced dataset, an intuitive motivation of active sampling methods is to modify the dataset with customized sampling mechanisms to get a balanced distribution. Random \textit{oversampling} and \textit{undersampling} are two main strategies. The mechanism of random \textit{oversampling} can be naturally implemented by adding a set of data from the minority class into the original dataset. Specifically, we randomly select $S$ samples from the minority class $D_{min}$ to duplicate and merge them with the original set. By doing so, the total number of minority class examples is augmented by $S$, leading to an adjustment in the class distribution balance. On the contrary, the random \textit{undersampling} method contributes to the class balance by reducing the number of samples in the majority class. In particular, we randomly select $S$ samples from the majority class $D_{maj}$ and remove them from the dataset. As a result, the overall samples are adjusted manually for a balanced distribution via both \textit{undersampling} and \textit{oversampling} strategies. 

\begin{figure}[htb]
    \centering
    \includegraphics[width=0.7\textwidth]{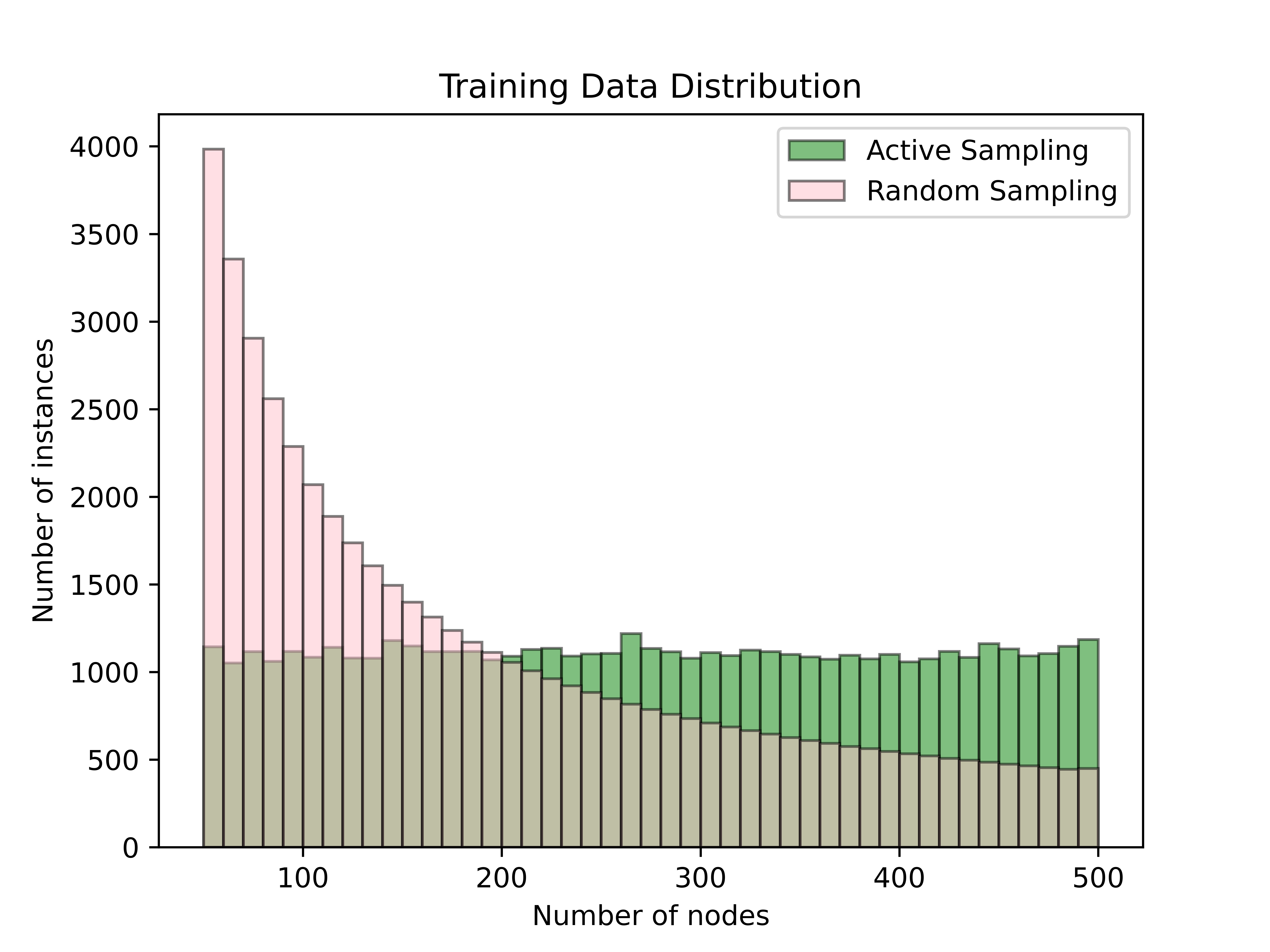}
    \caption{Training data distributions with different sampling strategies. The random sampling retains the original distribution of the training dataset, where small-scale data is the majority and large-scale data is the minority. The active sampling strategy can change the original distribution to a balanced distribution of the number of nodes.}
    \label{fig:distribution}
\end{figure}

In the link prediction task with scale-imbalanced TSP cases as the training data, we treat TSP instances with different numbers of cities as different classes. Hence for a TSP dataset with various sizes from 50 to 500 cities, there are 451 classes in total. Instead of removing or duplicating existing samples in the original dataset, we modify the traditional AS strategies by incorporating \textit{oversampling} and \textit{undersampling} into a \textit{uniform-sampling} on classes, with the aim to improve the sampling fairness and the data utilization. Specifically, we follow a two-step sampling manner for generating each batch of data during the training process. In the first step, we select $B$ classes uniformly from all classes of the number of cities, and $B$ is the batch size. In the second step, for each of the $B$ choices of the city number, we randomly sample an instance from the corresponding class, forming a batch of training data with a uniform distribution in the number of cities. In the original training process, all data from the training set are shuffled and divided into batches of data randomly. As a result, every batch of training data will follow a similar distribution to the overall dataset. On the contrary, if each batch of data is generated via a class-uniform sampling strategy, then the overall training data will follow a balanced distribution of the number of nodes, regardless of the original distribution of data. \textcolor{black}{The distribution of different training datasets obtained with different sampling strategies are shown in Figure \ref{fig:distribution}.}

\subsection{Dataset Generation}
\label{sec:dataset}

\begin{figure}
    \centering
    \includegraphics[width=1.0\textwidth]{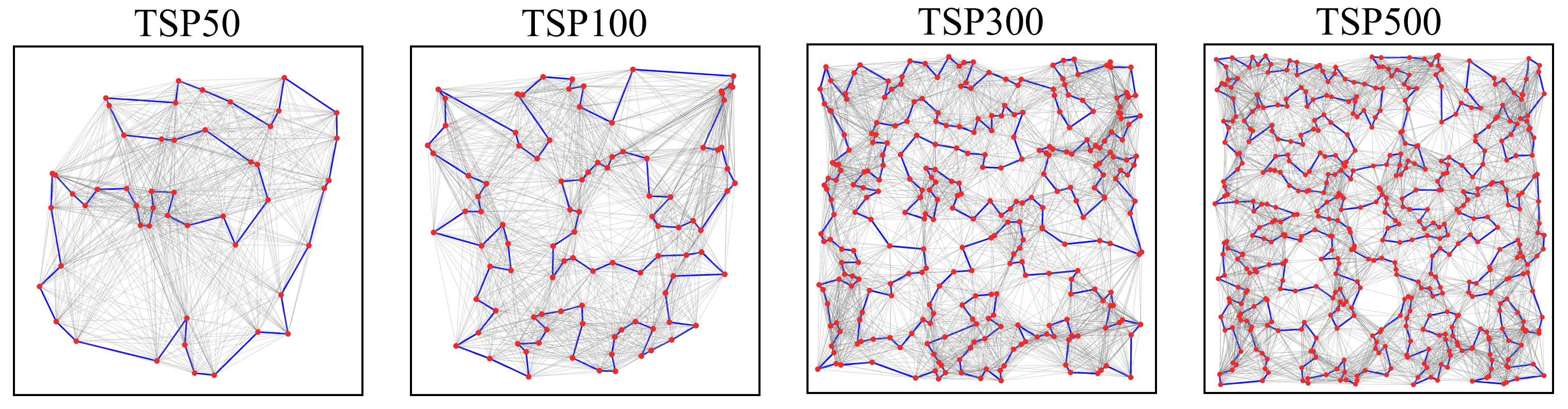}
    \caption{TSP instances of different scales in the training dataset. Blue edges indicate the optimal tour obtained by the Concorde solver.}
    \label{fig:optimal}
\end{figure}

To simulate the scale-imbalanced distribution in realistic optimization tasks, we create a TSP benchmark dataset with 50,000 instances as the training data. The dataset consists of graphs of different sizes from 50 to 500 nodes, and the number of graphs per size varies in inverse proportion to the number of nodes. As a result, there are 439 graphs with 50 nodes, and only 43 graphs have 500 nodes. Following the learning-based work in \citep{dwivedi2020benchmarking, joshi2019efficient, joshi2022learning}, the 2-dimensional locations $\left\{x,y\right\}$ of each node are uniformly sampled in the unit square ${[0,1]}^2$ for each instance. The optimal solution to each instance is provided by Concorde in the form of the ground truth label of each edge existing in the optimal tour. To test the generalization ability of models at various scales, we further create five separate test datasets for graphs of size $\left\{50, 100, 300, 500, 700\right\}$ nodes, each including 1,000 TSP instance-solution pairs. Figure \ref{fig:optimal} illustrates the optimal tours obtained by Concorde.

\section{Experimental Results}

In this section, we conduct systematic experiments on different amounts of training data, different scales of test cases, and different numbers of samples to compare the performance of our proposed approach with state-of-the-art learning-based NCO methods as well as the Concorde solver, in order to verify the effectiveness of the proposed method.

\subsection{Experimental Settings}
% 问题描述，不同训练集数量，不同测试集规模，不同采样规模（beam search和random search的选择），两种训练策略
As described in Section \ref{sec:overall_framework}, this work considers the 2-dimensional symmetric TSPs with $N$ cities randomly located in a unit square ${[0,1]}^2$ for simplification. The objective is to find an optimal route for a salesman to visit all cities once and return to the starting point, in order to minimize the total length of the tour. We formulate a TSP instance as a sparse graph representation based on heuristic information, where each node is a city and the edge weight represents the Euclidean distance between two cities. In the sparse graph representation, each node is only connected to its $K$ nearest neighbours according to the distances, where $K$ is a hyperparameter. The aim is to identify the shortest possible Hamiltonian cycle of the graph via link prediction and graph search methods. In our experimental setup, we configure four different levels of training data quantities with 1000, 5000, 10000, and 50000 instances. For each level of the training quantity, all instances are randomly sampled from the TSP benchmark dataset in Section \ref{sec:dataset}, forming a scale-imbalanced distribution where the number of instances is inversely proportional to the number of nodes. The test dataset comprises a total of 5000 TSP cases, including five distinct scales of instances: TSP50, TSP100, TSP300, TSP500, and TSP700, with each scale containing 1000 instances. It is worth noting that all test cases are completely independent of the training dataset. The optimal solutions to all test cases are generated by the Concorde solver. During the test process, the trained model outputs the link-predicted results for each input case in the form of a heatmap, indicating the probability of each edge existing in the optimal tour. Subsequently, the probability predictions are converted into feasible solutions via post-hoc graph search approaches. We explore two different post-hoc search strategies, namely beam search and random search, to generate feasible TSP solutions. Beam search is a fixed-width breadth-first searching approach that is widely adopted in generating high-probability sequences \citep{wu2016google}. Starting from the first node, the heatmap is explored recursively to construct $b$ solutions in parallel. Specifically, at each iteration, the top-$b$ edges with the highest probability in the heatmap will be selected and added to the $b$ partial tours. For each partial tour, the selected nodes will be masked to ensure the feasibility of the solution. The recursive selection process will continue to expand the $b$ tours until all nodes have been selected for each solution. The final prediction of beam search is determined by the tour with the minimum total length among the $b$ complete tours. Random search constructs a feasible solution to a TSP instance sequentially by adding one node at each step in an iterative manner. In contrast to the beam search where fixed $b$ nodes with the highest probabilities are selected at each step, random search determines the next node from all unvisited nodes via the roulette wheel selection, enabling the possibility of each node being selected with different probabilities. We adopt two search strategies in our experiments, and the results indicate that the random search achieves a better performance in terms of the solution quality. Therefore, we leverage the random search as our post-hoc graph search strategy in our experiments for all comparison methods, followed by a 2-opt local search to further enhance the model performance. We explore the effects of three different sampling quantities on solution performance by sampling 200, 500, and 1000 solutions for each test case. To investigate the effectiveness of the active sampling method introduced in Section \ref{sec:active_sampling}, we train all learning-based models in our comparative experiments using two training strategies separately, with and without the \textit{random active sampling} method.

\subsection{Baseline Methods}
We compare the proposed methodology with both an exact solver and state-of-the-art NCO methods as baselines. All learning-based models are trained in a supervised learning manner for a fair comparison.

\begin{enumerate}
    \item \textit{Concorde} \citep{applegate2007tsp}: Concorde is an exact solver specially designed for solving TSP. It builds on ideas from branch-and-cut algorithms, and incorporates linear programming and cutting plane methods to find optimal solutions to NP-hard combinatorial optimization problems including TSPs. Concorde has demonstrated exceptional performance in finding optimal solutions to TSP instances up to thousands of cities.

    % the Concorde solver is based on the branch-and-cut algorithm, which combines integer programming and cutting plane methods to solve challenging optimization problems.

    \item \textit{Multi-layer Perceptron (MLP)} \citep{hornik1989multilayer}: The MLP model takes the node features as inputs, with no consideration of the graph topology information. It first maps the node inputs to a high-dimensional latent space via multiple layers of linear projections, and generates edge embeddings by concatenating the two node embeddings of each edge. Finally, the edge embeddings are passed through a readout module to output probability predictions in the form of a heatmap.
    
    \item \textit{Graph Convolutional Network (GCN)} \citep{kipf2017semi}: The vanilla GCN model takes the graph-structured data as input, which incorporates both the node and edge features and the topology information. The message passing occurs between neighbour nodes via multiple layers of graph convolutional operators, where each node aggregates information about its neighbours with the same weights. 
    %The edge features are not involved in the message passing process, and node embeddings are concatenated together to form the edge embeddings.

    \item \textit{Graph Attention Network (GAT)} \citep{velivckovic2018graph}: The GAT model introduces masked self-attention layers and enables nodes to attend over the features of its neighbours by implicitly allocating different weights to neighbours during the message aggregation. For each graph attention layer in the GAT model, the input node features are first projected to higher-level expressions via a shared linear transformation, and then attention coefficients between each pair of nodes are calculated by a shared attention mechanism. The coefficients are normalized using the softmax function to serve as the weights of a linear combination of all neighbour embeddings for each node. The edge embeddings are represented as the concatenation of node embeddings.

    \item \textit{Residual Gated Graph ConvNets (GatedGCN)} \citep{bresson2017residual}: The GatedGCN model leverages an edge gating mechanism \citep{marcheggiani2017encoding} to allocate different weight vectors to neighbour node embeddings for the aggregation operator. It also incorporates the residual connections and batch normalization in the graph convolutional operation. Compared to the vanilla GCN model, the GatedGCN introduces explicit edge feature representation and update except for the message passing of node embeddings. The edge gates serve as a soft attention map with respect to the original sparse attention mechanism \citep{bahdanau2014neural, velivckovic2018graph}.
\end{enumerate}

\subsection{Implementation Details}
All learning-based models are trained with the same hyperparameter settings. For the model configuration, each model consists of $L=4$ message-passing layers with graph convolution-based operators, followed by an MLP block with 3 fully-connected layers to output link prediction values for all edges in the graph. The input node and edge features are 2 and 1 dimension, respectively, and the latent spaces of both node and edge embeddings have the same hidden dimension $H=64$ for each layer. We set $k=25$ in the $k$-nearest neighbour heuristic for all TSP instances of the training data. For the training process, we minimize the binary cross-entropy loss using the Adam optimizer, where the initial learning rate is $lr=0.001$ without decay for all models. The batch size is set to $B=32$ for all experiments, and the maximum number of training epochs is 500. During the test process, we evaluate the trained model on five test datasets of different problem scales: TSP50, TSP100, TSP300, TSP500, and TSP700 with a batch size of 100. The learning-based methods are run on a single NVIDIA A40 GPU with 48GB memory.

\subsection{Performance Indicators}
% F1, AUC, GAP
Following the previous work in \citep{joshi2019efficient, dwivedi2020benchmarking, joshi2022learning}, we evaluate the model performance on solving TSP cases via three indicators: F1 score, ROC AUC score, and the optimal gap.

\subsubsection{F1 score}
It is commonly adopted as the performance indicator for binary classification tasks. It is particularly preferable over accuracy when dealing with uneven datasets such as the TSP, where one class significantly dominates the other. For example, in a complete graph with $N^2$ edges, only the $2N$ edges in the optimal tour have positive labels. In such cases, the accuracy alone may not represent the model performance precisely. For a binary classification task, the numbers of true positive ($\mathrm{TP}$), false positive ($\mathrm{FP}$), and false negative ($\mathrm{FN}$) samples are used to calculate the following   $\mathrm{F1}$ score:
\setlength{\abovedisplayskip}{10pt}
 \begin{equation}
     \mathrm{F1}=\frac{\mathrm{TP}}{\mathrm{TP}+\frac{1}{2}(\mathrm{FP+FN})}.
 \end{equation}

 \subsubsection{ROC AUC score}
 It is defined as the area under the receiver operating characteristic curve from prediction scores. The $\mathrm{ROC\ AUC}$ score represents the effectiveness of a model in distinguishing between the positive and negative classes. A higher $\mathrm{ROC\ AUC}$ score implies a better performance at the binary classification task. It is suitable for class-imbalanced datasets, as it is not affected by changes in the classification threshold. For a binary classification task, $\mathrm{TPR}$ and $\mathrm{FPR}$ are the true positive rate and false positive rate for different threshold values. The $\mathrm{ROC\ AUC}$ score is calculated as:
 \begin{equation}
     \mathrm{ROC\ AUC}=\int_{t=0}^{1}\mathrm{TPR}(\mathrm{FPR}^{-1}(t))\,d t.
 \end{equation}
%https://en.wikipedia.org/wiki/Receiver_operating_characteristic#Area_under_the_curve
\subsubsection{Optimal gap}
It measures how close an obtained solution to the optimal one is. In our experiments, we define the optimal gap with respect to the Concorde solver, which calculates the normalized difference between the predicted tour length $S$ and the optimal solution $S_{best}$ provided by Concorde:
\begin{equation}
    \mathrm{GAP}=\frac{S-S_{best}}{S_{best}} \times 100\%.
\end{equation}

\begin{figure*}
    \centering
    \includegraphics[width=1.0\textwidth]{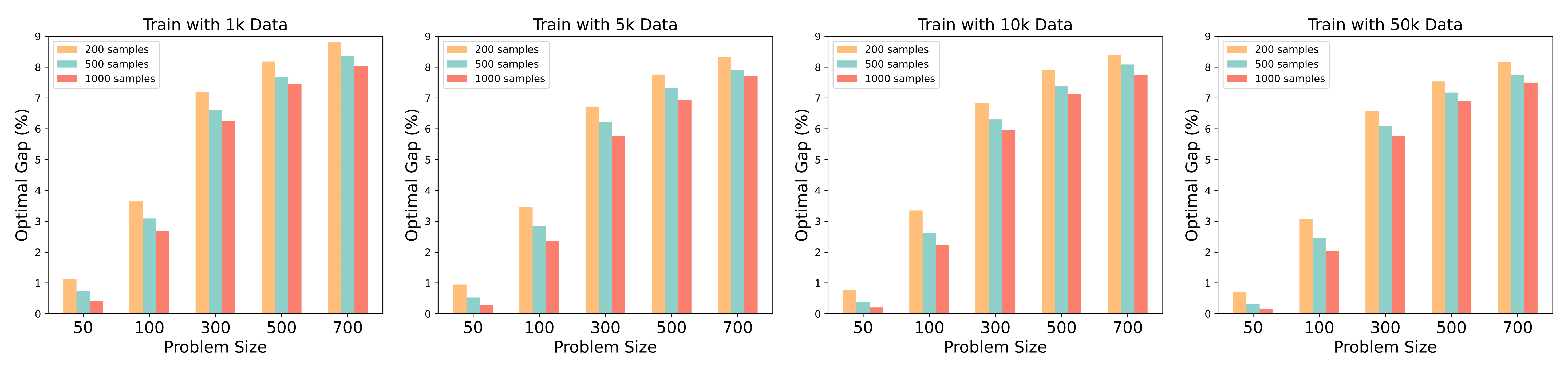}
    \caption{The optimal gap of different sample sizes on test cases.}
    \label{fig:sample_size}
\end{figure*}

\subsection{Results and Discussion}

%============opt GAP==============
\newgeometry{left=1.0cm,right=1.0cm,top=1.0cm,bottom=1.0cm}  % 调整页边距
\begin{table*}\tiny
\centering
\caption{\textcolor{black}{The average optimal gap (\%) on 1,000 test cases} w.r.t Concorde of learning-based models on TSP test datasets of different scales.}
\label{tab:gap}
\begin{tabular}{@{}ll|lll|lll|lll|lll|lll@{}}
\toprule
                                      & \multicolumn{1}{c|}{\textcolor{black}{Test Dataset}}            & \multicolumn{3}{c|}{TSP50}                                                    & \multicolumn{3}{c|}{TSP100}                                                   & \multicolumn{3}{c|}{TSP300}                                                   & \multicolumn{3}{c|}{TSP500}                                                   & \multicolumn{3}{c}{TSP700}                                                   \\ \midrule
                                      & \multicolumn{1}{c|}{Sample size} & \multicolumn{1}{c}{200} & \multicolumn{1}{c}{500} & \multicolumn{1}{c|}{1000} & \multicolumn{1}{c}{200} & \multicolumn{1}{c}{500} & \multicolumn{1}{c|}{1000} & \multicolumn{1}{c}{200} & \multicolumn{1}{c}{500} & \multicolumn{1}{c|}{1000} & \multicolumn{1}{c}{200} & \multicolumn{1}{c}{500} & \multicolumn{1}{c|}{1000} & \multicolumn{1}{c}{200} & \multicolumn{1}{c}{500} & \multicolumn{1}{c}{1000} \\ \midrule
\multirow{12}{*}{\rotatebox{90}{Train with 1k Data}}  & MLP                              & 1.854                   & 1.180                    & 0.731                     & 4.718                   & 4.047                   & 3.517                     & 8.226                   & 7.711                   & 7.362                     & 9.473                   & 8.979                   & 8.635                     & 10.003                  & 9.575                   & 9.310                     \\
                                      & GCN                              & 1.798                   & 1.083                   & 0.731                     & 4.761                   & 3.935                   & 3.484                     & 8.297                   & 7.666                   & 7.293                     & 9.215                   & 8.737                   & 8.400                       & 9.830                    & 9.382                   & 9.084                    \\
                                      & GAT                              & 1.955                   & 1.234                   & 0.796                     & 4.584                   & 3.867                   & 3.494                     & 8.188                   & 7.479                   & 7.168                     & 9.158                   & 8.641                   & 8.278                     & 9.625                   & 9.289                   & 8.99                     \\
                                      & GatedGCN                              & 1.627                   & 1.079                   & 0.774                     & 4.473                   & 3.737                   & 3.227                     & 7.973                   & 7.523                   & 7.124                     & 9.161                   & 8.717                   & 8.290                      & 9.703                   & 9.244                   & 9.042                    \\
                                      & GatedGCN*                             & 1.397                   & 0.872                   & 0.542                     & 4.053                   & 3.484                   & 3.006                     & 7.498                   & 7.079                   & 6.686                     & 8.692                   & 8.153                   & 7.769                     & 9.214                   & 8.818                   & 8.564                    \\ \cmidrule(l){2-17} 
                                      & MLP+AS                           & 1.743                   & 1.113                   & 0.819                     & 4.613                   & 3.857                   & 3.309                     & 7.966                   & 7.497                   & 7.182                     & 9.260                    & 8.809                   & 8.461                     & 10.002                  & 9.456                   & 9.117                    \\
                                      & GCN+AS                           & 1.859                   & 1.233                   & 0.888                     & 4.680                    & 3.933                   & 3.447                     & 8.338                   & 7.711                   & 7.342                     & 9.227                   & 8.708                   & 8.436                     & 9.867                   & 9.307                   & 9.077                    \\
                                      & GAT+AS                           & 1.826                   & 1.168                   & 0.811                     & 4.685                   & 3.945                   & 3.458                     & 7.910                    & 7.443                   & 7.093                     & 9.081                   & 8.522                   & 8.298                     & 9.664                   & 9.283                   & 8.965                    \\
                                      & GatedGCN+AS                           & 1.764                   & 0.960                    & 0.686                     & 4.420                    & 3.526                   & 3.065                     & 8.049                   & 7.457                   & 7.103                     & 9.089                   & 8.696                   & 8.420                      & 9.764                   & 9.398                   & 9.116                    \\
                                      & $\text{GatedGCN}^*$+AS                          & 1.261                   & 0.723                   & 0.434                     & 3.835                   & \textbf{3.084}          & \textbf{2.620}            & \textbf{7.136}          & 6.662                   & 6.324                     & 8.181                   & 7.793                   & 7.533                     & 8.720                    & 8.376                   & 8.079                    \\ \cmidrule(l){2-17} 
                                      & EdgeGAE (ours)                    & 1.148                   & \textbf{0.642}          & 0.435                     & 3.893                   & 3.111                   & 2.660                     & 7.149                   & \textbf{6.487}          & \textbf{6.100}            & \textbf{8.010}          & \textbf{7.646}          & \textbf{7.375}            & \textbf{8.629}          & \textbf{8.289}          & 8.032                    \\
                                      & EdgeGAE+AS (ours)                 & \textbf{1.120}          & 0.735                   & \textbf{0.424}            & \textbf{3.652}          & 3.091                   & 2.680                      & 7.182                   & 6.612                   & 6.250                      & 8.176                   & 7.669                   & 7.451                     & 8.796                   & 8.347                   & \textbf{8.029}           \\ \midrule \midrule
\multirow{12}{*}{\rotatebox{90}{Train with 5k Data}}  & MLP                              & 1.768                   & 1.134                   & 0.834                     & 4.603                   & 3.719                   & 3.251                     & 8.181                   & 7.669                   & 7.184                     & 9.402                   & 8.887                   & 8.528                     & 9.916                   & 9.499                   & 9.254                    \\
                                      & GCN                              & 1.904                   & 1.229                   & 0.859                     & 4.687                   & 3.909                   & 3.487                     & 8.066                   & 7.577                   & 7.262                     & 9.101                   & 8.728                   & 8.295                     & 9.531                   & 9.156                   & 8.827                    \\
                                      & GAT                              & 1.823                   & 1.069                   & 0.773                     & 4.677                   & 4.061                   & 3.420                     & 7.990                   & 7.486                   & 7.142                     & 8.933                   & 8.483                   & 8.072                     & 9.494                   & 9.145                   & 8.830                    \\
                                      & GatedGCN                              & 1.388                   & 0.781                   & 0.519                     & 3.725                   & 3.221                   & 2.727                     & 7.313                   & 6.829                   & 6.475                     & 8.383                   & 7.873                   & 7.648                     & 9.088                   & 8.712                   & 8.506                    \\
                                      & $\text{GatedGCN}^*$                             & 1.083                   & 0.609                   & 0.385                     & 3.890                   & 3.096                   & 2.641                     & 7.066                   & 6.576                   & 6.215                     & 7.997                   & 7.617                   & 7.272                     & 8.547                   & 8.194                   & 7.886                    \\ \cmidrule(l){2-17} 
                                      & MLP+AS                           & 1.764                   & 1.109                   & 0.704                     & 4.509                   & 3.760                    & 3.221                     & 7.972                   & 7.450                    & 7.075                     & 9.323                   & 8.848                   & 8.481                     & 9.886                   & 9.464                   & 9.199                    \\
                                      & GCN+AS                           & 1.772                   & 1.041                   & 0.645                     & 4.573                   & 3.857                   & 3.303                     & 8.040                   & 7.532                   & 7.092                     & 9.002                   & 8.371                   & 8.122                     & 9.424                   & 9.115                   & 8.790                     \\
                                      & GAT+AS                           & 1.788                   & 1.151                   & 0.802                     & 4.556                   & 3.871                   & 3.389                     & 7.980                   & 7.443                   & 7.113                     & 8.951                   & 8.490                   & 8.200                     & 9.470                   & 9.072                   & 8.811                    \\
                                      & GatedGCN+AS                           & 1.435                   & 0.901                   & 0.638                     & 4.361                   & 3.642                   & 3.065                     & 7.844                   & 7.291                   & 6.926                     & 8.888                   & 8.398                   & 8.055                     & 9.453                   & 9.114                   & 8.810                     \\
                                      & $\text{GatedGCN}^*$+AS                          & 1.190                   & 0.568                   & 0.315                     & 3.719                   & 3.033                   & 2.576                     & 7.113                   & 6.513                   & 6.208                     & 8.195                   & 7.668                   & 7.437                     & 8.816                   & 8.426                   & 8.184                    \\ \cmidrule(l){2-17} 
                                      & EdgeGAE (ours)                    & 0.958                   & 0.539                   & \textbf{0.280}            & 3.568                   & \textbf{2.768}          & 2.382                     & 6.864                   & 6.425                   & 5.992                     & 8.100                   & 7.562                   & 7.224                     & 8.438                   & 8.108                   & 7.837                    \\
                                      & EdgeGAE+AS (ours)                 & \textbf{0.949}          & \textbf{0.525}          & \textbf{0.280}            & \textbf{3.466}          & 2.856                   & \textbf{2.354}            & \textbf{6.719}          & \textbf{6.220}          & \textbf{5.767}            & \textbf{7.759}          & \textbf{7.325}          & \textbf{6.939}            & \textbf{8.318}          & \textbf{7.908}          & \textbf{7.697}           \\ \midrule \midrule
\multirow{12}{*}{\rotatebox{90}{Train with 10k Data}} & MLP                              & 1.933                   & 1.208                   & 0.811                     & 4.572                   & 3.783                   & 3.223                     & 8.188                   & 7.635                   & 7.284                     & 9.454                   & 9.008                   & 8.632                     & 9.934                   & 9.519                   & 9.203                    \\
                                      & GCN                              & 1.761                   & 1.123                   & 0.723                     & 4.721                   & 3.957                   & 3.490                      & 7.908                   & 7.287                   & 6.951                     & 9.004                   & 8.503                   & 8.140                     & 9.349                   & 8.946                   & 8.732                    \\
                                      & GAT                              & 1.930                   & 1.268                   & 0.805                     & 4.677                   & 3.979                   & 3.526                     & 7.917                   & 7.325                   & 7.019                     & 8.963                   & 8.536                   & 8.241                     & 9.433                   & 9.098                   & 8.766                    \\
                                      & GatedGCN                              & 1.544                   & 0.923                   & 0.589                     & 4.281                   & 3.482                   & 2.975                     & 7.572                   & 6.933                   & 6.667                     & 8.451                   & 7.944                   & 7.713                     & 9.030                   & 8.640                   & 8.370                    \\
                                      & $\text{GatedGCN}^*$                             & 1.283                   & 0.601                   & 0.405                     & 3.953                   & 3.206                   & 2.684                     & 7.371                   & 6.826                   & 6.401                     & 8.416                   & 7.929                   & 7.624                     & 9.114                   & 8.662                   & 8.398                    \\ \cmidrule(l){2-17} 
                                      & MLP+AS                           & 1.811                   & 1.071                   & 0.719                     & 4.569                   & 3.828                   & 3.337                     & 8.118                   & 7.522                   & 7.214                     & 9.280                   & 8.897                   & 8.543                     & 10.024                  & 9.544                   & 9.235                    \\
                                      & GCN+AS                           & 2.039                   & 1.253                   & 0.772                     & 4.536                   & 3.868                   & 3.327                     & 8.002                   & 7.500                   & 7.007                     & 8.996                   & 8.490                   & 8.131                     & 9.521                   & 9.037                   & 8.810                     \\
                                      & GAT+AS                           & 1.821                   & 1.150                   & 0.821                     & 4.502                   & 3.868                   & 3.369                     & 7.939                   & 7.328                   & 6.956                     & 8.868                   & 8.366                   & 8.164                     & 9.372                   & 8.996                   & 8.789                    \\
                                      & GatedGCN+AS                           & 1.518                   & 0.882                   & 0.586                     & 4.417                   & 3.743                   & 3.186                     & 7.744                   & 7.105                   & 6.772                     & 8.716                   & 8.306                   & 8.005                     & 9.273                   & 8.977                   & 8.697                    \\
                                      & $\text{GatedGCN}^*$+AS                          & 1.092                   & 0.582                   & 0.271                     & 3.742                   & 3.046                   & 2.632                     & 7.223                   & 6.566                   & 6.239                     & 8.012                   & 7.626                   & 7.260                     & 8.642                   & 8.287                   & 8.014                    \\ \cmidrule(l){2-17} 
                                      & EdgeGAE (ours)                    & \textbf{0.753}          & 0.366                   & 0.247                     & \textbf{3.253}          & 2.638                   & 2.257                     & \textbf{6.760}          & \textbf{6.246}          & 6.009                     & \textbf{7.821}          & 7.437                   & \textbf{7.074}            & \textbf{8.369}          & \textbf{7.971}          & \textbf{7.718}           \\
                                      & EdgeGAE+AS (ours)                 & 0.769                   & \textbf{0.365}          & \textbf{0.208}            & 3.352                   & \textbf{2.626}          & \textbf{2.233}            & 6.825                   & 6.301                   & \textbf{5.946}            & 7.897                   & \textbf{7.375}          & 7.128                     & 8.392                   & 8.082                   & 7.748                    \\ \midrule \midrule
\multirow{12}{*}{\rotatebox{90}{Train with 50k Data}} & MLP                              & 1.623                   & 1.052                   & 0.729                     & 4.461                   & 3.805                   & 3.325                     & 8.102                   & 7.634                   & 7.182                     & 9.198                   & 8.841                   & 8.500                     & 9.919                   & 9.513                   & 9.215                    \\
                                      & GCN                              & 1.858                   & 1.107                   & 0.787                     & 4.603                   & 3.930                    & 3.438                     & 7.920                   & 7.390                   & 7.033                     & 8.876                   & 8.405                   & 8.050                     & 9.342                   & 8.971                   & 8.696                    \\
                                      & GAT                              & 1.701                   & 1.174                   & 0.802                     & 4.507                   & 3.751                   & 3.288                     & 7.889                   & 7.379                   & 6.916                     & 8.799                   & 8.439                   & 8.078                     & 9.405                   & 8.951                   & 8.744                    \\
                                      & GatedGCN                              & 1.285                   & 0.759                   & 0.484                     & 4.318                   & 3.555                   & 3.010                     & 7.532                   & 7.098                   & 6.646                     & 8.503                   & 8.077                   & 7.753                     & 9.007                   & 8.69                    & 8.378                    \\
                                      & $\text{GatedGCN}^*$                             & 0.822                   & 0.415                   & 0.243                     & 3.417                   & 2.785                   & 2.373                     & 6.627                   & 6.196                   & 5.804                     & 7.835                   & 7.405                   & 7.053                     & 8.430                   & 8.072                   & 7.795                    \\ \cmidrule(l){2-17} 
                                      & MLP+AS                           & 1.621                   & 0.938                   & 0.678                     & 4.510                    & 3.860                    & 3.184                     & 8.131                   & 7.618                   & 7.086                     & 9.208                   & 8.800                   & 8.495                     & 9.874                   & 9.481                   & 9.220                    \\
                                      & GCN+AS                           & 1.891                   & 1.126                   & 0.727                     & 4.618                   & 3.876                   & 3.514                     & 7.889                   & 7.274                   & 6.985                     & 8.808                   & 8.350                   & 8.062                     & 9.275                   & 8.961                   & 8.737                    \\
                                      & GAT+AS                           & 1.727                   & 1.098                   & 0.769                     & 4.684                   & 3.830                    & 3.468                     & 8.042                   & 7.499                   & 7.061                     & 9.025                   & 8.543                   & 8.230                      & 9.457                   & 9.110                   & 8.831                    \\
                                      & GatedGCN+AS                           & 1.281                   & 0.694                   & 0.521                     & 4.124                   & 3.375                   & 3.002                     & 7.530                    & 6.951                   & 6.654                     & 8.462                   & 8.113                   & 7.811                     & 8.972                   & 8.667                   & 8.347                    \\
                                      & $\text{GatedGCN}^*$+AS                          & 0.886                   & 0.486                   & 0.247                     & 3.463                   & 2.812                   & 2.303                     & 7.052                   & 6.483                   & 6.068                     & 7.904                   & 7.400                   & 7.106                     & 8.459                   & 8.105                   & 7.857                    \\ \cmidrule(l){2-17} 
                                      & EdgeGAE (ours)                    & 0.723                   & \textbf{0.311}          & \textbf{0.137}            & \textbf{3.025}          & \textbf{2.330}           & 2.056                     & 6.652                   & \textbf{6.071}          & \textbf{5.640}            & 7.564                   & \textbf{7.112}          & \textbf{6.814}            & \textbf{8.026}          & 7.785                   & 7.502                    \\
                                      & EdgeGAE+AS (ours)                 & \textbf{0.694}          & 0.323                   & 0.168                     & 3.069                   & 2.463                   & \textbf{2.028}            & \textbf{6.572}          & 6.089                   & 5.772                     & \textbf{7.529}          & 7.170                    & 6.904                     & 8.161                   & \textbf{7.754}          & \textbf{7.499}           \\ \bottomrule
\end{tabular}
\raggedright
\begin{tablenotes}
    \item The mark $({}^*)$ indicates the usage of explicit edge representations in the message passing process. 
    \item The mark $(\text{+AS})$ indicates training the model with the \textit{random active sampling} strategy in Section \ref{sec:active_sampling}.
\end{tablenotes}
\end{table*}
\restoregeometry  % 恢复页边距

\subsubsection{Comparison against baseline models}
The performance of the proposed EdgeGAE versus the baseline methods in terms of the optimal gap is shown in Table \ref{tab:gap}. There are four sets of experiments with different amounts of training data. For each set of experiments, the results of test cases of different sizes are evaluated on models trained on the same set of data. The proposed EdgeGAE shows superior performance compared to other learning-based methods in most cases of the training data amounts and problem sizes. In the experiment with only 1000 training instances, EdgeGAE achieves the lowest optimal gap on TSP50, TSP500, and TSP700, regardless of the sample size. On the TSP100 test dataset, the proposed model exhibits a slightly inferior optimal gap compared to the GatedGCN model with explicit edge representations when the sample size of each case is set to 500 and 1000. One possible reason is that the small amount of training data is insufficient for the models to learn enough implicit knowledge, resulting in insignificant differences in performance between models at small and medium-scale problems. Among all the baseline methods, the MLP has the highest optimal gap, as it is the only learning-based model that solely takes node features as its input without considering any topology information. GCN performs worst among all the message passing-based models, because there is no attention mechanism which can allocate learnable weights to different neighbours during the message-passing process. In various scales of problem sizes, the GatedGCN with explicit edge representations exhibits significant performance improvement compared to the original version, revealing the importance of effective edge embeddings in link prediction tasks. In other words, merely considering node embeddings in message passing is insufficient for the model to generalize well on problems of various scales. In the experiment with 5000, 10000, and 50000 training instances, both the EdgeGAE models trained with and without \textit{random active sampling} strategy outperform other baseline methods at all settings of problem scales and sample sizes. The advantage becomes more notable as the number of training samples increases. The comparative results in terms of the optimal gap imply the effectiveness of the combination of the residual gated graph encoder and the edge-centered decoder model in link prediction for TSP tasks. Across all four experiment sets, the learning-based approaches exhibit a decline in performance with the increase of the problem sizes. The trained models achieve the highest average performance on TSP50 and the lowest on TSP700 test cases.

\subsubsection{Sensitivity analysis of sample sizes}
To investigate the influence of different sample sizes on the model performance, we sample 200, 500, and 1000 solutions for each test case in each group of experiments. Figure \ref{fig:sample_size} compares the average optimal gaps of EdgeGAE with different sample sizes under four experimental settings. It can be observed that a higher number of samples leads to a lower gap across all scales of test cases. It stands to reason that increasing the number of sampled solutions for a particular TSP instance increases the likelihood of obtaining better objective values. Nonetheless, it is worth noting that a greater number of samples generally comes with additional computational overhead. Consequently, it is crucial in real-world applications to balance the model performance and computational costs. Furthermore, the comparative results also show that the average optimal gap increases as the problem scale grows, which is irrelevant to the sample size.

\subsubsection{Generalization ability}
To demonstrate the ability of EdgeGAE to generalize over larger instances than it has ever seen, we evaluate the model performance on TSP700 cases after training on imbalanced data ranging from 50 to 500 cities. In Table \ref{tab:gap}, the optimal gap on TSP700 serves as the performance indicator of the proposed model and compared baselines. It can be observed that given a sufficient number of samples per case, the proposed model achieves an average optimal gap on TSP700 which is remarkably close to that of TSP500. Compared to other baseline methods, the proposed EdgeGAE exhibits a superior ability to generalize to large-scale problems. Furthermore, the model trained on a larger amount of small-scale instances tends to generalize better on large scales than those with limited training data. Hence, it can be concluded that the model can benefit from more training data even sampled from small-scale instances.

\subsubsection{Inference time and complexity analysis}
\begin{figure}
    \centering
    \includegraphics[width=0.5\textwidth]{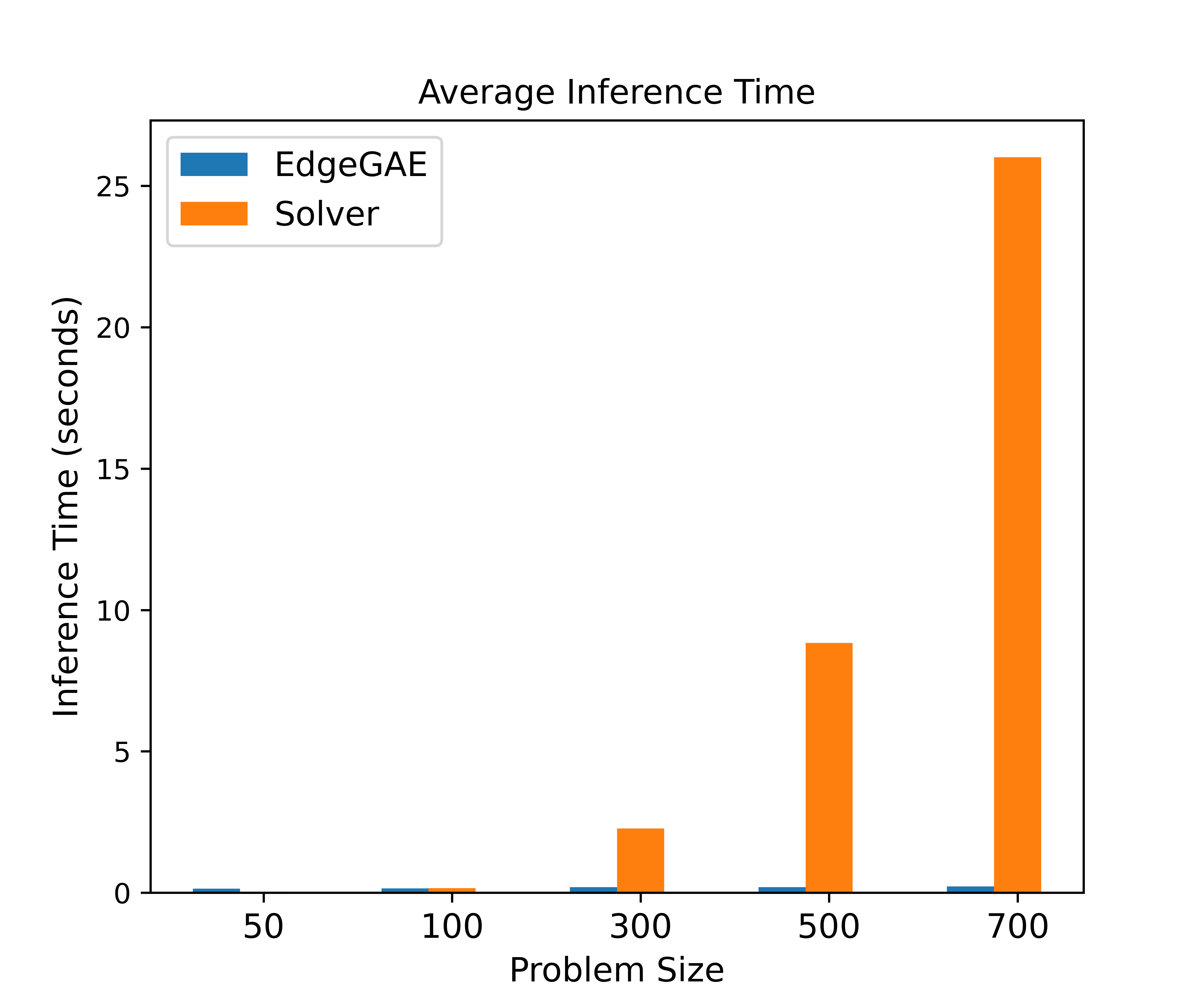}
    \caption{Average inference time of the EdgeGAE and the classical TSP solver Concorde on TSP instances of different scales.}
    \label{fig:inference}
\end{figure}

We further compare the inference time of the proposed EdgeGAE and the classical TSP solver on different scales of TSP instances, \textcolor{black}{and the statistical results are presented in Figure \ref{fig:inference}}. For each problem size, the trained EdgeGAE model followed by graph search makes prediction on 1000 unseen TSP instances to obtain feasible solutions. The Concorde solver performs optimization on the same set of problems for a fair comparison. Figure 7 illustrates the average inference times of two methods for each test case at five different scales. It is worth noting that on smaller-sized cases with no more than 100 cities, both the EdgeGAE and the Concorde solver require extremely short inference times with a negligible difference. However, as the problem size increases, the inference time taken by the solver to find an optimal solution grows exponentially. By contrast, the inference time of EdgeGAE remains basically constant, which is not affected by the problem size. This result implies that the EdgeGAE has a significant advantage over traditional solvers in terms of inference time, especially for solving large-scale optimization problems with real-time decision requirements. \textcolor{black}{In the EdgeGAE model, the message passing process for nodes is calculated according to Equation \ref{eq:resgated_node} with the time complexity of $O(d_n \cdot |\mathcal{N}|)$, where $d_n$ is the dimension of the node representation, and $|\mathcal{N}|$ is the size of the neighborhood. Similarly, the message passing of edges in Equation \ref{eq:resgated_edge} has the time complexity of $(d_e^2)$, where $d_e$ is the dimension of the edge representation. Therefore, the overall time complexity for each message passing layer in EdgeGAE is $O(d_n \cdot |\mathcal{N}_i| + d_e^2)$.}

\subsection{Ablation Study}
% could also have a barplot here
To investigate the effect of different sampling strategies during the training process on model performance, we conduct an ablation study to train each comparison model on the same set of training data using random shuffle sampling and random active sampling strategies separately, and evaluate them on different sizes of test cases. The model performances are compared in Table \ref{tab:gap}, where each subset of models without the mark $(\text{+AS})$ is trained with randomly shuffled data batches, and each subset of models with the mark $(\text{+AS})$ is trained with batches of data generated via the random active sampling strategy in Section \ref{sec:active_sampling}. For a fair comparison, both sampling strategies share the same set of training data in each set of settings. The optimal gaps of EdgeGAE present noteworthy findings. When the training dataset is limited to only 1,000 instances, the performance of random shuffle sampling surpasses that of active sampling. One reasonable explanation is that the active sampling from an extremely small dataset causes the model to focus solely on a portion of the data, thereby intensifying the model overfitting on training data. When the training dataset contains 5,000 scale-imbalanced cases, the active sampling strategy outperforms random shuffle sampling significantly in any case of problem scales and sample sizes. The advantages of active sampling over random shuffle sampling become less noticeable as the quantity of training data continues to increase. These results indicate that having a considerable amount of training data can partially mitigate the effect of scale-imbalanced distributions on model performance, advocating the necessity of using active sampling strategies on small-scale constrained datasets with size-skewed distributions.

\section{Conclusion}

In this paper, we investigate solving scale-imbalanced TSPs as link predictions on graphs. To cope with this task, we propose a data-driven NCO framework named EdgeGAE, which enhances the original encoder-decoder framework with a residual gated encoder model and an edge-centered decoder model. The residual gated encoder model introduces explicit edge embeddings in the message-passing process for link prediction tasks. It incorporates residual connections and an edge gated attention mechanism into the vanilla graph convolution. The edge-centered decoder model replaces the inner product with an edge-aware decoding strategy, where each edge aggregates information from its source and target nodes via an edge-centered message-passing scheme. The proposed EdgeGAE framework enables the input graph to capture topology information and learn latent representations via message passing on both nodes and edges for link prediction on TSPs. To address the scale-imbalanced data distribution issues, we further introduce an active sampling strategy for the training process. We generate a scale-imbalanced TSP benchmark dataset with 50,000 instances and compare the proposed framework with state-of-the-art learning-based methods. Experimental results on various problem scales show that EdgeGAE outperforms other baselines in terms of solution quality, sample efficiency, and generalization ability.

Despite the competitive performance of the proposed method, it should be noted that there is still a gap between the learning-based NCO methods and exact solvers. To further improve the model performance with limited training data, we will consider incorporating semi-supervised learning and reinforcement learning into the learning process of data-driven NCO approaches. Moreover, existing NCO models are mostly trained in an offline manner and require all data to be centralized in a local storage. Considering the increasing demand for privacy protection, our future work will extend data-driven NCO to federated learning settings where privacy data is kept locally without sharing during the learning process. We will also consider NCO with online optimization to deal with time-varying problems in real-world applications.

\section*{Acknowledgements}
This work was supported in part by the National Natural Science Foundation of China under Grant No. 62006053 and No.62136003.


\appendix

% \section{Sample Appendix Section}
% \label{sec:sample:appendix}
% Lorem ipsum dolor sit amet, consectetur adipiscing elit, sed do eiusmod tempor section \ref{sec:sample1} incididunt ut labore et dolore magna aliqua. Ut enim ad minim veniam, quis nostrud exercitation ullamco laboris nisi ut aliquip ex ea commodo consequat. Duis aute irure dolor in reprehenderit in voluptate velit esse cillum dolore eu fugiat nulla pariatur. Excepteur sint occaecat cupidatat non proident, sunt in culpa qui officia deserunt mollit anim id est laborum.

\bibliographystyle{elsarticle-num-names} 
\bibliography{cas-refs}

\begin{thebibliography}{67}
\expandafter\ifx\csname natexlab\endcsname\relax\def\natexlab#1{#1}\fi
\providecommand{\url}[1]{\texttt{#1}}
\providecommand{\href}[2]{#2}
\providecommand{\path}[1]{#1}
\providecommand{\DOIprefix}{doi:}
\providecommand{\ArXivprefix}{arXiv:}
\providecommand{\URLprefix}{URL: }
\providecommand{\Pubmedprefix}{pmid:}
\providecommand{\doi}[1]{\href{http://dx.doi.org/#1}{\path{#1}}}
\providecommand{\Pubmed}[1]{\href{pmid:#1}{\path{#1}}}
\providecommand{\bibinfo}[2]{#2}
\ifx\xfnm\relax \def\xfnm[#1]{\unskip,\space#1}\fi
%Type = Article
\bibitem[{Bengio et~al.(2021)Bengio, Lodi, and Prouvost}]{bengio2021machine}
\bibinfo{author}{Y.~Bengio}, \bibinfo{author}{A.~Lodi}, \bibinfo{author}{A.~Prouvost},
\newblock \bibinfo{title}{Machine learning for combinatorial optimization: a methodological tour d’horizon},
\newblock \bibinfo{journal}{European Journal of Operational Research} \bibinfo{volume}{290} (\bibinfo{year}{2021}) \bibinfo{pages}{405--421}.
%Type = Article
\bibitem[{Xu et~al.(2022)Xu, Fang, Chen, Xu, Du, and Zhang}]{xu2022reinforcement}
\bibinfo{author}{Y.~Xu}, \bibinfo{author}{M.~Fang}, \bibinfo{author}{L.~Chen}, \bibinfo{author}{G.~Xu}, \bibinfo{author}{Y.~Du}, \bibinfo{author}{C.~Zhang},
\newblock \bibinfo{title}{Reinforcement learning with multiple relational attention for solving vehicle routing problems},
\newblock \bibinfo{journal}{IEEE Transactions on Cybernetics} \bibinfo{volume}{52} (\bibinfo{year}{2022}) \bibinfo{pages}{11107--11120}.
%Type = Inproceedings
\bibitem[{Sanches et~al.(2017)Sanches, Whitley, and Tin{\'o}s}]{sanches2017improving}
\bibinfo{author}{D.~Sanches}, \bibinfo{author}{D.~Whitley}, \bibinfo{author}{R.~Tin{\'o}s},
\newblock \bibinfo{title}{Improving an exact solver for the traveling salesman problem using partition crossover},
\newblock in: \bibinfo{booktitle}{Proceedings of the Genetic and Evolutionary Computation Conference}, \bibinfo{year}{2017}, pp. \bibinfo{pages}{337--344}.
%Type = Article
\bibitem[{Rosenkrantz et~al.(1977)Rosenkrantz, Stearns, and Lewis}]{rosenkrantz1977analysis}
\bibinfo{author}{D.~J. Rosenkrantz}, \bibinfo{author}{R.~E. Stearns}, \bibinfo{author}{P.~M. Lewis, II},
\newblock \bibinfo{title}{An analysis of several heuristics for the traveling salesman problem},
\newblock \bibinfo{journal}{SIAM journal on computing} \bibinfo{volume}{6} (\bibinfo{year}{1977}) \bibinfo{pages}{563--581}.
%Type = Article
\bibitem[{Helsgaun(2017)}]{helsgaun2017extension}
\bibinfo{author}{K.~Helsgaun},
\newblock \bibinfo{title}{An extension of the lin-kernighan-helsgaun tsp solver for constrained traveling salesman and vehicle routing problems}  (\bibinfo{year}{2017}).
%Type = Article
\bibitem[{Cui et~al.(2024)Cui, Wu, Huang, Xu, Liu, and Xiao}]{cui2024multi}
\bibinfo{author}{J.~Cui}, \bibinfo{author}{L.~Wu}, \bibinfo{author}{X.~Huang}, \bibinfo{author}{D.~Xu}, \bibinfo{author}{C.~Liu}, \bibinfo{author}{W.~Xiao},
\newblock \bibinfo{title}{Multi-strategy adaptable ant colony optimization algorithm and its application in robot path planning},
\newblock \bibinfo{journal}{Knowledge-Based Systems}  (\bibinfo{year}{2024}) \bibinfo{pages}{111459}.
%Type = Article
\bibitem[{Wang et~al.(2023{\natexlab{a}})Wang, Ding, and Jin}]{wang2023multi}
\bibinfo{author}{S.~Wang}, \bibinfo{author}{B.~Ding}, \bibinfo{author}{Y.~Jin},
\newblock \bibinfo{title}{A multi-factorial evolutionary algorithm with asynchronous optimization processes for solving the robust influence maximization problem},
\newblock \bibinfo{journal}{IEEE Computational Intelligence Magazine} \bibinfo{volume}{18} (\bibinfo{year}{2023}{\natexlab{a}}) \bibinfo{pages}{41--53}.
%Type = Article
\bibitem[{Wang et~al.(2023{\natexlab{b}})Wang, Jin, and Cai}]{wang2023enhancing}
\bibinfo{author}{S.~Wang}, \bibinfo{author}{Y.~Jin}, \bibinfo{author}{M.~Cai},
\newblock \bibinfo{title}{Enhancing the robustness of networks against multiple damage models using a multifactorial evolutionary algorithm},
\newblock \bibinfo{journal}{IEEE Transactions on Systems, Man, and Cybernetics: Systems}  (\bibinfo{year}{2023}{\natexlab{b}}).
%Type = Article
\bibitem[{Wang and Tan(2022)}]{wang2022solving}
\bibinfo{author}{S.~Wang}, \bibinfo{author}{X.~Tan},
\newblock \bibinfo{title}{Solving the robust influence maximization problem on multi-layer networks via a memetic algorithm},
\newblock \bibinfo{journal}{Applied Soft Computing} \bibinfo{volume}{121} (\bibinfo{year}{2022}) \bibinfo{pages}{108750}.
%Type = Inproceedings
\bibitem[{Bello et~al.(2017)Bello, Pham, Le, Norouzi, and Bengio}]{bello2017neural}
\bibinfo{author}{I.~Bello}, \bibinfo{author}{H.~Pham}, \bibinfo{author}{Q.~V. Le}, \bibinfo{author}{M.~Norouzi}, \bibinfo{author}{S.~Bengio},
\newblock \bibinfo{title}{Neural combinatorial optimization with reinforcement learning},
\newblock in: \bibinfo{booktitle}{International Conference on Learning Representations}, \bibinfo{year}{2017}.
%Type = Article
\bibitem[{Jin et~al.(2018)Jin, Wang, Chugh, Guo, and Miettinen}]{jin2018data}
\bibinfo{author}{Y.~Jin}, \bibinfo{author}{H.~Wang}, \bibinfo{author}{T.~Chugh}, \bibinfo{author}{D.~Guo}, \bibinfo{author}{K.~Miettinen},
\newblock \bibinfo{title}{Data-driven evolutionary optimization: An overview and case studies},
\newblock \bibinfo{journal}{IEEE Transactions on Evolutionary Computation} \bibinfo{volume}{23} (\bibinfo{year}{2018}) \bibinfo{pages}{442--458}.
%Type = Book
\bibitem[{Jin et~al.(2021)Jin, Wang, and Sun}]{jin2021data}
\bibinfo{author}{Y.~Jin}, \bibinfo{author}{H.~Wang}, \bibinfo{author}{C.~Sun}, \bibinfo{title}{Data-driven evolutionary optimization}, \bibinfo{publisher}{Springer}, \bibinfo{year}{2021}.
%Type = Article
\bibitem[{Hopfield and Tank(1985)}]{hopfield1985neural}
\bibinfo{author}{J.~J. Hopfield}, \bibinfo{author}{D.~W. Tank},
\newblock \bibinfo{title}{“neural” computation of decisions in optimization problems},
\newblock \bibinfo{journal}{Biological cybernetics} \bibinfo{volume}{52} (\bibinfo{year}{1985}) \bibinfo{pages}{141--152}.
%Type = Article
\bibitem[{Angeniol et~al.(1988)Angeniol, Vaubois, and Le~Texier}]{angeniol1988self}
\bibinfo{author}{B.~Angeniol}, \bibinfo{author}{G.~D. L.~C. Vaubois}, \bibinfo{author}{J.-Y. Le~Texier},
\newblock \bibinfo{title}{Self-organizing feature maps and the travelling salesman problem},
\newblock \bibinfo{journal}{Neural Networks} \bibinfo{volume}{1} (\bibinfo{year}{1988}) \bibinfo{pages}{289--293}.
%Type = Article
\bibitem[{Sutskever et~al.(2014)Sutskever, Vinyals, and Le}]{sutskever2014sequence}
\bibinfo{author}{I.~Sutskever}, \bibinfo{author}{O.~Vinyals}, \bibinfo{author}{Q.~V. Le},
\newblock \bibinfo{title}{Sequence to sequence learning with neural networks},
\newblock \bibinfo{journal}{Advances in Neural Information Processing Systems} \bibinfo{volume}{27} (\bibinfo{year}{2014}).
%Type = Article
\bibitem[{Wu et~al.(2016)Wu, Schuster, Chen, Le, Norouzi, Macherey, Krikun, Cao, Gao, Macherey et~al.}]{wu2016google}
\bibinfo{author}{Y.~Wu}, \bibinfo{author}{M.~Schuster}, \bibinfo{author}{Z.~Chen}, \bibinfo{author}{Q.~V. Le}, \bibinfo{author}{M.~Norouzi}, \bibinfo{author}{W.~Macherey}, \bibinfo{author}{M.~Krikun}, \bibinfo{author}{Y.~Cao}, \bibinfo{author}{Q.~Gao}, \bibinfo{author}{K.~Macherey}, et~al.,
\newblock \bibinfo{title}{Google's neural machine translation system: Bridging the gap between human and machine translation},
\newblock \bibinfo{journal}{arXiv preprint arXiv:1609.08144}  (\bibinfo{year}{2016}).
%Type = Article
\bibitem[{Vinyals et~al.(2015)Vinyals, Fortunato, and Jaitly}]{vinyals2015pointer}
\bibinfo{author}{O.~Vinyals}, \bibinfo{author}{M.~Fortunato}, \bibinfo{author}{N.~Jaitly},
\newblock \bibinfo{title}{Pointer networks},
\newblock \bibinfo{journal}{Advances in Neural Information Processing Systems} \bibinfo{volume}{28} (\bibinfo{year}{2015}).
%Type = Article
\bibitem[{Bahdanau et~al.(2014)Bahdanau, Cho, and Bengio}]{bahdanau2014neural}
\bibinfo{author}{D.~Bahdanau}, \bibinfo{author}{K.~Cho}, \bibinfo{author}{Y.~Bengio},
\newblock \bibinfo{title}{Neural machine translation by jointly learning to align and translate},
\newblock \bibinfo{journal}{arXiv preprint arXiv:1409.0473}  (\bibinfo{year}{2014}).
%Type = Article
\bibitem[{Chen et~al.(2020)Chen, Wang, Wang, and Kuo}]{chen2020graph}
\bibinfo{author}{F.~Chen}, \bibinfo{author}{Y.-C. Wang}, \bibinfo{author}{B.~Wang}, \bibinfo{author}{C.-C.~J. Kuo},
\newblock \bibinfo{title}{Graph representation learning: a survey},
\newblock \bibinfo{journal}{APSIPA Transactions on Signal and Information Processing} \bibinfo{volume}{9} (\bibinfo{year}{2020}) \bibinfo{pages}{e15}.
%Type = Article
\bibitem[{Zhao and Cheong(2023)}]{zhao2023obfuscating}
\bibinfo{author}{J.~Zhao}, \bibinfo{author}{K.~H. Cheong},
\newblock \bibinfo{title}{Obfuscating community structure in complex network with evolutionary divide-and-conquer strategy},
\newblock \bibinfo{journal}{IEEE Transactions on Evolutionary Computation}  (\bibinfo{year}{2023}).
%Type = Article
\bibitem[{Zhao et~al.(2023)Zhao, Wang, Cao, and Cheong}]{zhao2023self}
\bibinfo{author}{J.~Zhao}, \bibinfo{author}{Z.~Wang}, \bibinfo{author}{J.~Cao}, \bibinfo{author}{K.~H. Cheong},
\newblock \bibinfo{title}{A self-adaptive evolutionary deception framework for community structure},
\newblock \bibinfo{journal}{IEEE Transactions on Systems, Man, and Cybernetics: Systems}  (\bibinfo{year}{2023}).
%Type = Article
\bibitem[{Joshi et~al.(2019)Joshi, Laurent, and Bresson}]{joshi2019efficient}
\bibinfo{author}{C.~K. Joshi}, \bibinfo{author}{T.~Laurent}, \bibinfo{author}{X.~Bresson},
\newblock \bibinfo{title}{An efficient graph convolutional network technique for the travelling salesman problem},
\newblock \bibinfo{journal}{arXiv preprint arXiv:1906.01227}  (\bibinfo{year}{2019}).
%Type = Inproceedings
\bibitem[{Kipf and Welling(2017)}]{kipf2017semi}
\bibinfo{author}{T.~N. Kipf}, \bibinfo{author}{M.~Welling},
\newblock \bibinfo{title}{Semi-supervised classification with graph convolutional networks},
\newblock in: \bibinfo{booktitle}{International Conference on Learning Representations}, \bibinfo{year}{2017}.
%Type = Inproceedings
\bibitem[{Veli{\v{c}}kovi{\'c} et~al.(2018)Veli{\v{c}}kovi{\'c}, Cucurull, Casanova, Romero, Li{\`o}, and Bengio}]{velivckovic2018graph}
\bibinfo{author}{P.~Veli{\v{c}}kovi{\'c}}, \bibinfo{author}{G.~Cucurull}, \bibinfo{author}{A.~Casanova}, \bibinfo{author}{A.~Romero}, \bibinfo{author}{P.~Li{\`o}}, \bibinfo{author}{Y.~Bengio},
\newblock \bibinfo{title}{Graph attention networks},
\newblock in: \bibinfo{booktitle}{International Conference on Learning Representations}, \bibinfo{year}{2018}.
%Type = Article
\bibitem[{Xu et~al.(2018)Xu, Hu, Leskovec, and Jegelka}]{xu2018powerful}
\bibinfo{author}{K.~Xu}, \bibinfo{author}{W.~Hu}, \bibinfo{author}{J.~Leskovec}, \bibinfo{author}{S.~Jegelka},
\newblock \bibinfo{title}{How powerful are graph neural networks?},
\newblock \bibinfo{journal}{arXiv preprint arXiv:1810.00826}  (\bibinfo{year}{2018}).
%Type = Inproceedings
\bibitem[{Nowak et~al.(2018)Nowak, Villar, Bandeira, and Bruna}]{nowak2018revised}
\bibinfo{author}{A.~Nowak}, \bibinfo{author}{S.~Villar}, \bibinfo{author}{A.~S. Bandeira}, \bibinfo{author}{J.~Bruna},
\newblock \bibinfo{title}{Revised note on learning quadratic assignment with graph neural networks},
\newblock in: \bibinfo{booktitle}{2018 IEEE Data Science Workshop}, \bibinfo{organization}{IEEE}, \bibinfo{year}{2018}, pp. \bibinfo{pages}{1--5}.
%Type = Article
\bibitem[{Joshi et~al.(2022)Joshi, Cappart, Rousseau, and Laurent}]{joshi2022learning}
\bibinfo{author}{C.~K. Joshi}, \bibinfo{author}{Q.~Cappart}, \bibinfo{author}{L.-M. Rousseau}, \bibinfo{author}{T.~Laurent},
\newblock \bibinfo{title}{Learning the travelling salesperson problem requires rethinking generalization},
\newblock \bibinfo{journal}{Constraints} \bibinfo{volume}{27} (\bibinfo{year}{2022}) \bibinfo{pages}{70--98}.
%Type = Inproceedings
\bibitem[{Liu et~al.(2023)Liu, Yan, and Jin}]{liu2023end}
\bibinfo{author}{S.~Liu}, \bibinfo{author}{X.~Yan}, \bibinfo{author}{Y.~Jin},
\newblock \bibinfo{title}{End-to-end pareto set prediction with graph neural networks for multi-objective facility location},
\newblock in: \bibinfo{booktitle}{Proceedings of the 12th International Conference on Evolutionary Multi-Criterion Optimization}, \bibinfo{organization}{Springer}, \bibinfo{year}{2023}, pp. \bibinfo{pages}{147--161}.
%Type = Article
\bibitem[{Wang et~al.(2023)Wang, Yan, and Jin}]{wang2023graph}
\bibinfo{author}{X.~Wang}, \bibinfo{author}{X.~Yan}, \bibinfo{author}{Y.~Jin},
\newblock \bibinfo{title}{A graph neural network with negative message passing for graph coloring},
\newblock \bibinfo{journal}{arXiv preprint arXiv:2301.11164}  (\bibinfo{year}{2023}).
%Type = Article
\bibitem[{Dai et~al.(2017)Dai, Khalil, Zhang, Dilkina, and Song}]{dai2017learning}
\bibinfo{author}{H.~Dai}, \bibinfo{author}{E.~Khalil}, \bibinfo{author}{Y.~Zhang}, \bibinfo{author}{B.~Dilkina}, \bibinfo{author}{L.~Song},
\newblock \bibinfo{title}{Learning combinatorial optimization algorithms over graphs},
\newblock \bibinfo{journal}{Advances in Neural Information Processing Systems} \bibinfo{volume}{30} (\bibinfo{year}{2017}).
%Type = Article
\bibitem[{Li et~al.(2018)Li, Chen, and Koltun}]{li2018combinatorial}
\bibinfo{author}{Z.~Li}, \bibinfo{author}{Q.~Chen}, \bibinfo{author}{V.~Koltun},
\newblock \bibinfo{title}{Combinatorial optimization with graph convolutional networks and guided tree search},
\newblock \bibinfo{journal}{Advances in neural information processing systems} \bibinfo{volume}{31} (\bibinfo{year}{2018}).
%Type = Inproceedings
\bibitem[{Deudon et~al.(2018)Deudon, Cournut, Lacoste, Adulyasak, and Rousseau}]{deudon2018learning}
\bibinfo{author}{M.~Deudon}, \bibinfo{author}{P.~Cournut}, \bibinfo{author}{A.~Lacoste}, \bibinfo{author}{Y.~Adulyasak}, \bibinfo{author}{L.-M. Rousseau},
\newblock \bibinfo{title}{Learning heuristics for the {TSP} by policy gradient},
\newblock in: \bibinfo{booktitle}{Integration of Constraint Programming, Artificial Intelligence, and Operations Research: 15th International Conference, CPAIOR 2018}, \bibinfo{organization}{Springer}, \bibinfo{year}{2018}, pp. \bibinfo{pages}{170--181}.
%Type = Inproceedings
\bibitem[{Kool et~al.(2019)Kool, van Hoof, and Welling}]{kool2019attention}
\bibinfo{author}{W.~Kool}, \bibinfo{author}{H.~van Hoof}, \bibinfo{author}{M.~Welling},
\newblock \bibinfo{title}{Attention, learn to solve routing problems!},
\newblock in: \bibinfo{booktitle}{International Conference on Learning Representations}, \bibinfo{year}{2019}.
%Type = Article
\bibitem[{Dwivedi et~al.(2020)Dwivedi, Joshi, Laurent, Bengio, and Bresson}]{dwivedi2020benchmarking}
\bibinfo{author}{V.~P. Dwivedi}, \bibinfo{author}{C.~K. Joshi}, \bibinfo{author}{T.~Laurent}, \bibinfo{author}{Y.~Bengio}, \bibinfo{author}{X.~Bresson},
\newblock \bibinfo{title}{Benchmarking graph neural networks},
\newblock \bibinfo{journal}{arXiv preprint arXiv:2003.00982}  (\bibinfo{year}{2020}).
%Type = Article
\bibitem[{Bresson and Laurent(2017)}]{bresson2017residual}
\bibinfo{author}{X.~Bresson}, \bibinfo{author}{T.~Laurent},
\newblock \bibinfo{title}{Residual gated graph convnets},
\newblock \bibinfo{journal}{arXiv preprint arXiv:1711.07553}  (\bibinfo{year}{2017}).
%Type = Inproceedings
\bibitem[{Fu et~al.(2021)Fu, Qiu, and Zha}]{fu2021generalize}
\bibinfo{author}{Z.-H. Fu}, \bibinfo{author}{K.-B. Qiu}, \bibinfo{author}{H.~Zha},
\newblock \bibinfo{title}{Generalize a small pre-trained model to arbitrarily large tsp instances},
\newblock in: \bibinfo{booktitle}{Proceedings of the AAAI Conference on Artificial Intelligence}, volume~\bibinfo{volume}{35}, \bibinfo{year}{2021}, pp. \bibinfo{pages}{7474--7482}.
%Type = Article
\bibitem[{Li et~al.(2023)Li, Ma, Cao, Wu, Song, Zhang, and Chee}]{li2023learning}
\bibinfo{author}{J.~Li}, \bibinfo{author}{Y.~Ma}, \bibinfo{author}{Z.~Cao}, \bibinfo{author}{Y.~Wu}, \bibinfo{author}{W.~Song}, \bibinfo{author}{J.~Zhang}, \bibinfo{author}{Y.~M. Chee},
\newblock \bibinfo{title}{Learning feature embedding refiner for solving vehicle routing problems},
\newblock \bibinfo{journal}{IEEE Transactions on Neural Networks and Learning Systems}  (\bibinfo{year}{2023}).
%Type = Inproceedings
\bibitem[{Luo et~al.(2023)Luo, Lin, Liu, Zhang, and Wang}]{luo2023neural}
\bibinfo{author}{F.~Luo}, \bibinfo{author}{X.~Lin}, \bibinfo{author}{F.~Liu}, \bibinfo{author}{Q.~Zhang}, \bibinfo{author}{Z.~Wang},
\newblock \bibinfo{title}{Neural combinatorial optimization with heavy decoder: Toward large scale generalization},
\newblock in: \bibinfo{booktitle}{Thirty-seventh Conference on Neural Information Processing Systems}, \bibinfo{year}{2023}.
%Type = Inproceedings
\bibitem[{Zhou et~al.(2023)Zhou, Wu, Song, Cao, and Zhang}]{zhou2023towards}
\bibinfo{author}{J.~Zhou}, \bibinfo{author}{Y.~Wu}, \bibinfo{author}{W.~Song}, \bibinfo{author}{Z.~Cao}, \bibinfo{author}{J.~Zhang},
\newblock \bibinfo{title}{Towards omni-generalizable neural methods for vehicle routing problems},
\newblock in: \bibinfo{booktitle}{the 40th International Conference on Machine Learning (ICML 2023)}, \bibinfo{year}{2023}.
%Type = Article
\bibitem[{Browne et~al.(2012)Browne, Powley, Whitehouse, Lucas, Cowling, Rohlfshagen, Tavener, Perez, Samothrakis, and Colton}]{browne2012survey}
\bibinfo{author}{C.~B. Browne}, \bibinfo{author}{E.~Powley}, \bibinfo{author}{D.~Whitehouse}, \bibinfo{author}{S.~M. Lucas}, \bibinfo{author}{P.~I. Cowling}, \bibinfo{author}{P.~Rohlfshagen}, \bibinfo{author}{S.~Tavener}, \bibinfo{author}{D.~Perez}, \bibinfo{author}{S.~Samothrakis}, \bibinfo{author}{S.~Colton},
\newblock \bibinfo{title}{A survey of monte carlo tree search methods},
\newblock \bibinfo{journal}{IEEE Transactions on Computational Intelligence and AI in games} \bibinfo{volume}{4} (\bibinfo{year}{2012}) \bibinfo{pages}{1--43}.
%Type = Inproceedings
\bibitem[{Zhang et~al.(2023)Zhang, Xiao, Wang, Song, and Chen}]{zhang2023neural}
\bibinfo{author}{D.~Zhang}, \bibinfo{author}{Z.~Xiao}, \bibinfo{author}{Y.~Wang}, \bibinfo{author}{M.~Song}, \bibinfo{author}{G.~Chen},
\newblock \bibinfo{title}{Neural tsp solver with progressive distillation},
\newblock in: \bibinfo{booktitle}{Proceedings of the AAAI Conference on Artificial Intelligence}, volume~\bibinfo{volume}{37}, \bibinfo{year}{2023}, pp. \bibinfo{pages}{12147--12154}.
%Type = Article
\bibitem[{Xin et~al.(2021)Xin, Song, Cao, and Zhang}]{xin2021neurolkh}
\bibinfo{author}{L.~Xin}, \bibinfo{author}{W.~Song}, \bibinfo{author}{Z.~Cao}, \bibinfo{author}{J.~Zhang},
\newblock \bibinfo{title}{Neurolkh: Combining deep learning model with lin-kernighan-helsgaun heuristic for solving the traveling salesman problem},
\newblock \bibinfo{journal}{Advances in Neural Information Processing Systems} \bibinfo{volume}{34} (\bibinfo{year}{2021}) \bibinfo{pages}{7472--7483}.
%Type = Inproceedings
\bibitem[{Zheng et~al.(2021)Zheng, He, Zhou, Jin, and Li}]{zheng2021combining}
\bibinfo{author}{J.~Zheng}, \bibinfo{author}{K.~He}, \bibinfo{author}{J.~Zhou}, \bibinfo{author}{Y.~Jin}, \bibinfo{author}{C.-M. Li},
\newblock \bibinfo{title}{Combining reinforcement learning with lin-kernighan-helsgaun algorithm for the traveling salesman problem},
\newblock in: \bibinfo{booktitle}{Proceedings of the AAAI conference on artificial intelligence}, volume~\bibinfo{volume}{35}, \bibinfo{year}{2021}, pp. \bibinfo{pages}{12445--12452}.
%Type = Article
\bibitem[{Colorni et~al.(1996)Colorni, Dorigo, Maffioli, Maniezzo, Righini, and Trubian}]{colorni1996heuristics}
\bibinfo{author}{A.~Colorni}, \bibinfo{author}{M.~Dorigo}, \bibinfo{author}{F.~Maffioli}, \bibinfo{author}{V.~Maniezzo}, \bibinfo{author}{G.~Righini}, \bibinfo{author}{M.~Trubian},
\newblock \bibinfo{title}{Heuristics from nature for hard combinatorial optimization problems},
\newblock \bibinfo{journal}{International Transactions in Operational Research} \bibinfo{volume}{3} (\bibinfo{year}{1996}) \bibinfo{pages}{1--21}.
%Type = Inproceedings
\bibitem[{Ye et~al.(2023)Ye, Wang, Cao, Liang, and Li}]{ye2023deepaco}
\bibinfo{author}{H.~Ye}, \bibinfo{author}{J.~Wang}, \bibinfo{author}{Z.~Cao}, \bibinfo{author}{H.~Liang}, \bibinfo{author}{Y.~Li},
\newblock \bibinfo{title}{Deepaco: Neural-enhanced ant systems for combinatorial optimization},
\newblock in: \bibinfo{booktitle}{Thirty-seventh Conference on Neural Information Processing Systems}, \bibinfo{year}{2023}.
%Type = Article
\bibitem[{Helsgaun(2000)}]{helsgaun2000effective}
\bibinfo{author}{K.~Helsgaun},
\newblock \bibinfo{title}{An effective implementation of the lin--kernighan traveling salesman heuristic},
\newblock \bibinfo{journal}{European journal of operational research} \bibinfo{volume}{126} (\bibinfo{year}{2000}) \bibinfo{pages}{106--130}.
%Type = Article
\bibitem[{Helsgaun(2009)}]{helsgaun2009general}
\bibinfo{author}{K.~Helsgaun},
\newblock \bibinfo{title}{General k-opt submoves for the lin--kernighan tsp heuristic},
\newblock \bibinfo{journal}{Mathematical Programming Computation} \bibinfo{volume}{1} (\bibinfo{year}{2009}) \bibinfo{pages}{119--163}.
%Type = Article
\bibitem[{Lin(1965)}]{lin1965computer}
\bibinfo{author}{S.~Lin},
\newblock \bibinfo{title}{Computer solutions of the traveling salesman problem},
\newblock \bibinfo{journal}{Bell System Technical Journal} \bibinfo{volume}{44} (\bibinfo{year}{1965}) \bibinfo{pages}{2245--2269}.
%Type = Article
\bibitem[{Watkins and Dayan(1992)}]{watkins1992q}
\bibinfo{author}{C.~J. Watkins}, \bibinfo{author}{P.~Dayan},
\newblock \bibinfo{title}{Q-learning},
\newblock \bibinfo{journal}{Machine learning} \bibinfo{volume}{8} (\bibinfo{year}{1992}) \bibinfo{pages}{279--292}.
%Type = Book
\bibitem[{Sutton and Barto(2018)}]{sutton2018reinforcement}
\bibinfo{author}{R.~S. Sutton}, \bibinfo{author}{A.~G. Barto}, \bibinfo{title}{Reinforcement learning: An introduction}, \bibinfo{publisher}{MIT press}, \bibinfo{year}{2018}.
%Type = Article
\bibitem[{Wu et~al.(2024)Wu, Wu, Wu, Feng, and Tan}]{wu2024evolutionary}
\bibinfo{author}{X.~Wu}, \bibinfo{author}{S.-h. Wu}, \bibinfo{author}{J.~Wu}, \bibinfo{author}{L.~Feng}, \bibinfo{author}{K.~C. Tan},
\newblock \bibinfo{title}{Evolutionary computation in the era of large language model: Survey and roadmap},
\newblock \bibinfo{journal}{arXiv preprint arXiv:2401.10034}  (\bibinfo{year}{2024}).
%Type = Article
\bibitem[{Liu et~al.(2024)Liu, Tong, Yuan, Lin, Luo, Wang, Lu, and Zhang}]{liu2024example}
\bibinfo{author}{F.~Liu}, \bibinfo{author}{X.~Tong}, \bibinfo{author}{M.~Yuan}, \bibinfo{author}{X.~Lin}, \bibinfo{author}{F.~Luo}, \bibinfo{author}{Z.~Wang}, \bibinfo{author}{Z.~Lu}, \bibinfo{author}{Q.~Zhang},
\newblock \bibinfo{title}{An example of evolutionary computation+ large language model beating human: Design of efficient guided local search},
\newblock \bibinfo{journal}{arXiv preprint arXiv:2401.02051}  (\bibinfo{year}{2024}).
%Type = Article
\bibitem[{Wang et~al.(2024)Wang, Yu, McAleer, Yu, and Yang}]{wang2024asp}
\bibinfo{author}{C.~Wang}, \bibinfo{author}{Z.~Yu}, \bibinfo{author}{S.~McAleer}, \bibinfo{author}{T.~Yu}, \bibinfo{author}{Y.~Yang},
\newblock \bibinfo{title}{Asp: Learn a universal neural solver!},
\newblock \bibinfo{journal}{IEEE Transactions on Pattern Analysis and Machine Intelligence}  (\bibinfo{year}{2024}).
%Type = Article
\bibitem[{Battaglia et~al.(2018)Battaglia, Hamrick, Bapst, Sanchez-Gonzalez, Zambaldi, Malinowski, Tacchetti, Raposo, Santoro, Faulkner et~al.}]{battaglia2018relational}
\bibinfo{author}{P.~W. Battaglia}, \bibinfo{author}{J.~B. Hamrick}, \bibinfo{author}{V.~Bapst}, \bibinfo{author}{A.~Sanchez-Gonzalez}, \bibinfo{author}{V.~Zambaldi}, \bibinfo{author}{M.~Malinowski}, \bibinfo{author}{A.~Tacchetti}, \bibinfo{author}{D.~Raposo}, \bibinfo{author}{A.~Santoro}, \bibinfo{author}{R.~Faulkner}, et~al.,
\newblock \bibinfo{title}{Relational inductive biases, deep learning, and graph networks},
\newblock \bibinfo{journal}{arXiv preprint arXiv:1806.01261}  (\bibinfo{year}{2018}).
%Type = Inproceedings
\bibitem[{Gilmer et~al.(2017)Gilmer, Schoenholz, Riley, Vinyals, and Dahl}]{gilmer2017neural}
\bibinfo{author}{J.~Gilmer}, \bibinfo{author}{S.~S. Schoenholz}, \bibinfo{author}{P.~F. Riley}, \bibinfo{author}{O.~Vinyals}, \bibinfo{author}{G.~E. Dahl},
\newblock \bibinfo{title}{Neural message passing for quantum chemistry},
\newblock in: \bibinfo{booktitle}{International conference on machine learning}, \bibinfo{organization}{PMLR}, \bibinfo{year}{2017}, pp. \bibinfo{pages}{1263--1272}.
%Type = Article
\bibitem[{Ma et~al.(2019)Ma, Ge, He, Thaker, and Drori}]{ma2019combinatorial}
\bibinfo{author}{Q.~Ma}, \bibinfo{author}{S.~Ge}, \bibinfo{author}{D.~He}, \bibinfo{author}{D.~Thaker}, \bibinfo{author}{I.~Drori},
\newblock \bibinfo{title}{Combinatorial optimization by graph pointer networks and hierarchical reinforcement learning},
\newblock \bibinfo{journal}{arXiv preprint arXiv:1911.04936}  (\bibinfo{year}{2019}).
%Type = Article
\bibitem[{Kwon et~al.(2020)Kwon, Choo, Kim, Yoon, Gwon, and Min}]{kwon2020pomo}
\bibinfo{author}{Y.-D. Kwon}, \bibinfo{author}{J.~Choo}, \bibinfo{author}{B.~Kim}, \bibinfo{author}{I.~Yoon}, \bibinfo{author}{Y.~Gwon}, \bibinfo{author}{S.~Min},
\newblock \bibinfo{title}{Pomo: Policy optimization with multiple optima for reinforcement learning},
\newblock \bibinfo{journal}{Advances in Neural Information Processing Systems} \bibinfo{volume}{33} (\bibinfo{year}{2020}) \bibinfo{pages}{21188--21198}.
%Type = Inproceedings
\bibitem[{Kool et~al.(2022)Kool, van Hoof, Gromicho, and Welling}]{kool2022deep}
\bibinfo{author}{W.~Kool}, \bibinfo{author}{H.~van Hoof}, \bibinfo{author}{J.~Gromicho}, \bibinfo{author}{M.~Welling},
\newblock \bibinfo{title}{Deep policy dynamic programming for vehicle routing problems},
\newblock in: \bibinfo{booktitle}{Integration of Constraint Programming, Artificial Intelligence, and Operations Research: 19th International Conference, CPAIOR 2022}, \bibinfo{organization}{Springer}, \bibinfo{year}{2022}, pp. \bibinfo{pages}{190--213}.
%Type = Article
\bibitem[{Freitag and Al-Onaizan(2017)}]{freitag2017beam}
\bibinfo{author}{M.~Freitag}, \bibinfo{author}{Y.~Al-Onaizan},
\newblock \bibinfo{title}{Beam search strategies for neural machine translation},
\newblock \bibinfo{journal}{arXiv preprint arXiv:1702.01806}  (\bibinfo{year}{2017}).
%Type = Inproceedings
\bibitem[{Huber and Raidl(2022)}]{huber2022learning}
\bibinfo{author}{M.~Huber}, \bibinfo{author}{G.~R. Raidl},
\newblock \bibinfo{title}{Learning beam search: Utilizing machine learning to guide beam search for solving combinatorial optimization problems},
\newblock in: \bibinfo{booktitle}{Machine Learning, Optimization, and Data Science: 7th International Conference}, \bibinfo{organization}{Springer}, \bibinfo{year}{2022}, pp. \bibinfo{pages}{283--298}.
%Type = Article
\bibitem[{Croes(1958)}]{croes1958method}
\bibinfo{author}{G.~A. Croes},
\newblock \bibinfo{title}{A method for solving traveling-salesman problems},
\newblock \bibinfo{journal}{Operations research} \bibinfo{volume}{6} (\bibinfo{year}{1958}) \bibinfo{pages}{791--812}.
%Type = Article
\bibitem[{Kipf and Welling(2016)}]{kipf2016variational}
\bibinfo{author}{T.~N. Kipf}, \bibinfo{author}{M.~Welling},
\newblock \bibinfo{title}{Variational graph auto-encoders},
\newblock \bibinfo{journal}{Advances in Neural Information Processing Systems, Bayesian Deep Learning Workshop}  (\bibinfo{year}{2016}).
%Type = Article
\bibitem[{Sen et~al.(2008)Sen, Namata, Bilgic, Getoor, Galligher, and Eliassi-Rad}]{sen2008collective}
\bibinfo{author}{P.~Sen}, \bibinfo{author}{G.~Namata}, \bibinfo{author}{M.~Bilgic}, \bibinfo{author}{L.~Getoor}, \bibinfo{author}{B.~Galligher}, \bibinfo{author}{T.~Eliassi-Rad},
\newblock \bibinfo{title}{Collective classification in network data},
\newblock \bibinfo{journal}{AI magazine} \bibinfo{volume}{29} (\bibinfo{year}{2008}) \bibinfo{pages}{93--93}.
%Type = Article
\bibitem[{He and Garcia(2009)}]{he2009learning}
\bibinfo{author}{H.~He}, \bibinfo{author}{E.~A. Garcia},
\newblock \bibinfo{title}{Learning from imbalanced data},
\newblock \bibinfo{journal}{IEEE Transactions on knowledge and data engineering} \bibinfo{volume}{21} (\bibinfo{year}{2009}) \bibinfo{pages}{1263--1284}.
%Type = Book
\bibitem[{Applegate et~al.(2007)Applegate, Bixby, Chvátal, and Cook}]{applegate2007tsp}
\bibinfo{author}{D.~L. Applegate}, \bibinfo{author}{R.~E. Bixby}, \bibinfo{author}{V.~Chvátal}, \bibinfo{author}{W.~J. Cook}, \bibinfo{year}{2007}.
%Type = Article
\bibitem[{Hornik et~al.(1989)Hornik, Stinchcombe, and White}]{hornik1989multilayer}
\bibinfo{author}{K.~Hornik}, \bibinfo{author}{M.~Stinchcombe}, \bibinfo{author}{H.~White},
\newblock \bibinfo{title}{Multilayer feedforward networks are universal approximators},
\newblock \bibinfo{journal}{Neural networks} \bibinfo{volume}{2} (\bibinfo{year}{1989}) \bibinfo{pages}{359--366}.
%Type = Inproceedings
\bibitem[{Marcheggiani and Titov(2017)}]{marcheggiani2017encoding}
\bibinfo{author}{D.~Marcheggiani}, \bibinfo{author}{I.~Titov},
\newblock \bibinfo{title}{Encoding sentences with graph convolutional networks for semantic role labeling},
\newblock in: \bibinfo{booktitle}{EMNLP 2017: Conference on Empirical Methods in Natural Language Processing}, \bibinfo{organization}{Association for Computational Linguistics}, \bibinfo{year}{2017}, pp. \bibinfo{pages}{1506--1515}.

\end{thebibliography}





% \end{thebibliography}
\end{document}